\pdfoutput=1
\documentclass[10pt,twocolumn,letterpaper]{article}

\usepackage[pagenumbers]{cvpr}
\makeatletter
\patchcmd{\@maketitle}{\vskip .375in}{\vspace*{-8.5pt}}{}{\PatchFailed}
\patchcmd{\@maketitle}{\vspace*{12pt}}{\vspace*{7pt}}{}{\PatchFailed}
\patchcmd{\@maketitle}{\vspace*{24pt}}{\vspace*{17pt}}{}{\PatchFailed}
\patchcmd{\maketitlesupplementary}{Supplementary Material}{Appendix}{}{\PatchFailed}
\makeatother

\usepackage{cuted}

\usepackage{currfile}

\usepackage{caption}

\usepackage{colortbl}
\definecolor{tablebestcolor}{HTML}{D1E7D8}
\definecolor{tablesecondcolor}{HTML}{E9F3DA}
\definecolor{tablerefgray}{HTML}{EDEDED}
\newcommand{\tablebest}[1]{\cellcolor{tablebestcolor}\textbf{#1}}
\newcommand{\tablesecond}[1]{\cellcolor{tablesecondcolor}\underline{#1}}
\DeclareRobustCommand{\legendbest}[1]{{\setlength{\fboxsep}{1.5pt}\colorbox{tablebestcolor}{\textbf{#1}}}}
\DeclareRobustCommand{\legendsecond}[1]{{\setlength{\fboxsep}{1.5pt}\colorbox{tablesecondcolor}{\underline{#1}}}}
\newcommand{\tablestd}[1]{{\mdseries\scriptsize\,$\pm$\,#1}}
\usepackage{pifont}

\usepackage[percent]{overpic}

\usepackage[skins,breakable]{tcolorbox}
\usepackage{fvextra}
\fvset{breaklines=true,breaksymbolleft={},fontsize=\fontsize{7}{8.4}\selectfont}
\newtcolorbox{promptbox}[1]{enhanced,breakable,colback=white,colframe=black!40,colbacktitle=black!7,coltitle=black,fonttitle=\footnotesize\bfseries,title={#1},boxrule=0.4pt,arc=1.5pt,left=4pt,right=4pt,top=2pt,bottom=2pt,toptitle=1.5pt,bottomtitle=1.5pt,before skip=6pt,after skip=6pt}
\usepackage{algorithm}
\usepackage[noend]{algpseudocode}

\definecolor{cvprblue}{rgb}{0.21,0.49,0.74}
\usepackage[pagebackref,breaklinks,colorlinks,allcolors=cvprblue]{hyperref}

\def\confName{CVPR}
\def\confYear{2027}

\usepackage{bm}

\DeclareMathOperator*{\argmin}{argmin}

\newcommand{\NICKNAME}{\textsc{Mundo}}

\title{\NICKNAME: Towards Customizable World Models}

\author{
Yihang Luo$^{1}$ \quad Zunnan Xu$^{1}$ \quad Shangchen Zhou$^{2}$ \quad Yushi Lan$^{3}$ \quad Kang Liao$^{1}$\\
Zerui Gong$^{4}$ \quad Xiao-Ming Wu$^{1}$ \quad Qingyi Tao$^{4}$ \quad Zhonghua Wu$^{4}$ \quad Bixuan Jin$^{4}$\\
Fei Wang$^{4}$ \quad Zhi Hou$^{4}$ \quad Xiaogang Wang$^{4}$ \quad Xingang Pan$^{1}$ \quad Chen Change Loy$^{1}$\\[1mm]
\normalsize $^{1}$S-Lab, Nanyang Technological University \quad $^{2}$Imperial College London \quad $^{3}$University of Oxford \quad $^{4}$ACE Robotics\\[-1pt]
\normalsize \url{https://mundo-world-model.github.io/}
}

\begin{document}
\maketitle
\begin{strip}
\vspace{-4em}
\centering
\includegraphics[width=\textwidth]{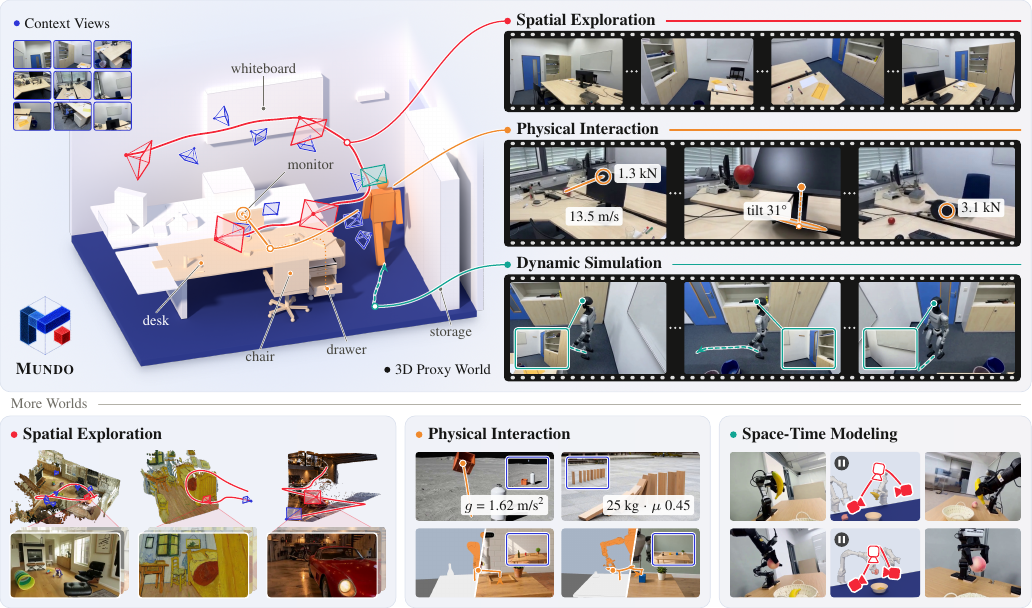}
\captionsetup{hypcap=false}
\captionof{figure}{\textbf{\NICKNAME{} is a customizable world model that turns context views of a scene into a world to explore, interact with, and simulate.} \textbf{Top:} Its explicit 3D proxy world is edited and simulated in code and rendered by a video model with the appearance of the context views: simulation handles physics, and the video model focuses on generation. \textbf{Bottom:} Under camera control, \NICKNAME{} explores worlds built from photos, a painting, and a generated image, edits physical properties such as gravity and friction, and renders bullet time.}
\captionsetup{hypcap=true}
\label{fig:teaser}%
\vspace{-8pt}
\end{strip}

\begin{abstract}

World models simulate how environments evolve under actions and physical conditions.
Existing video models capture rich appearance and motion priors, but implicit scene states limit direct control over the objects and physical properties that determine an interaction.
We present \NICKNAME, a customizable world model that represents objects and their physical properties in an editable, executable scene program.
Given context views of a scene, a vision-language agent recovers the object layout and refines it through automatic visual verification and numerical fitting.
The program separates spatial layout and physical properties from appearance, allowing a coarse geometric proxy to support interaction while the images retain scene detail.
Executing the program computes motion and contact under specified actions, forces, and physical settings.
The resulting states are rendered as proxy frames that guide a video model, with the context views providing appearance references.
\NICKNAME{} supports spatial exploration and controllable physical simulation, while outperforming existing methods in 3D proxy reconstruction and camera-controlled video generation.

\end{abstract}

\vspace{-3mm}

\section{Introduction}
\label{sec:intro}

World models represent the state and dynamics of a world, including its objects, spatial relationships, physical properties, and responses to actions~\cite{ha2018worldmodels,hafner2023dreamerv3}. In robotics and content creation, these capabilities often need to be instantiated for a specific environment~\cite{torne2024rialto,bruce2024genie}, usually with just a few images. The created world should then allow an agent to navigate the space, manipulate its objects, and simulate how changes in physical conditions affect their motion. Throughout these interactions, the simulated world should remain faithful to the observed environment in both geometry and appearance, while its dynamics should follow the specified actions and physical conditions. Achieving this requires jointly modeling scene appearance, spatial layout, actions and physical properties that may not have been observed together during training.

Video-based world models generate future observations conditioned on visual history and actions~\cite{bruce2024genie,alonso2024diamond,valevski2024gamengen}. Recent methods improve scene consistency across viewpoints~\cite{yu2024viewcrafter,ren2025gen3c} and preserve motion from source videos~\cite{lin2026vista4d,lan2026holo4d}. These advances, however, primarily improve visual fidelity and temporal consistency; the physical properties governing motion, such as mass and friction, remain implicit and difficult to control. Consequently, visually realistic predictions are not necessarily physically valid~\cite{li2025pisa}. Moreover, reproducing an observed trajectory captures what happened under one set of conditions, but does not determine how the motion should change under a different force or friction setting. Supporting such interventions requires an explicit, executable representation of the objects, their physical properties, and the relationships governing their motion.

Programs and simulators provide such explicit representations through editable states and executable rules~\cite{chen2026cwm,huang2026pwm,foo2026psivg,li2025wonderplay,liu2026realwonder,liang2026codevideomodel}, allowing motion to be recomputed when actions or physical properties change. Building such an executable world for a customized environment, however, first requires recovering its objects and spatial relationships from images. Real-to-sim methods estimate this structure~\cite{chen2024urdformer,torne2024rialto,mandi2024real2code,chen2025physgen3d,huang2026literealityagent}, but their reconstructed scenes often remain visually different from the source. Closing this appearance gap with conventional rendering requires detailed geometry and materials, even though physical interaction may need only object proxies without detailed appearance.

We believe these limitations can be addressed by composing specialized capabilities at inference time, namely perception, reasoning, simulation, and video generation, rather than relying on a single end-to-end model to predict the complete outcome.
This design separates the executable world state, which holds the object layout and physical properties, from the visual detail retained in the images.
Object proxies therefore need only encode the geometry required for interaction, and scenes, actions, and physical settings can be recombined without having appeared together in training.
The system can also benefit from stronger components, as we show for the agent model in \cref{sec:exp_ablation}.

We present \NICKNAME, a customizable world model that constructs an executable environment from context views. To bridge perception, reasoning, simulation, and video generation, we introduce two explicit interfaces: an editable scene program connects perception and reasoning to simulation, while rendered proxy frames connect simulation to video generation. 
To build the scene program, a vision-language agent~\cite{openai2026gpt6astra,anthropic2026claudeopus55} composes the proposals of a 3D detector~\cite{detone2026boxer} into one scene by reasoning over all context views, deciding which objects to keep, merge, remove, or add. It then refines the scene by comparing proxy renders with the context views and correcting inaccurate boxes with numerical fitting tools~\cite{mazur2026agentstar,huang2026literealityagent}. A physics engine~\cite{todorov2012mujoco} executes the program under the specified actions and physical settings, and a video model~\cite{minimax2026h3}, conditioned on the resulting proxy frames~\cite{chen2026cwm} and the context views, renders the simulated states with the appearance of the captured scene.

We evaluate \NICKNAME{} on CA-1M~\cite{lazarow2025cubify} and ADT~\cite{pan2023adt} for 3D proxy reconstruction, ScanNet++~\cite{yeshwanth2023scannetpp} and Tanks and Temples~\cite{knapitsch2017tanks} for camera-controlled video generation, and a physical interaction benchmark for physics-grounded video generation.
On CA-1M, given the same inputs, \NICKNAME{} improves the F1 at IoU 0.25 of the strongest baseline, Boxer~\cite{detone2026boxer}, from 41.0 to 51.3.
In camera-controlled video generation, it achieves the highest PSNR among the compared methods on both ScanNet++ and Tanks and Temples.
In physics-grounded video generation, its videos follow the simulated trajectories most faithfully, with a trajectory IoU of 0.83.
Our key contributions are as follows:
\begin{itemize}
\setlength{\itemsep}{0pt}
\item A customizable world model that grounds an editable, executable world in the context views of a real scene, so that the scene can be explored and physically interacted with while preserving its appearance.
\item An agentic reconstruction in which a vision-language agent composes detector proposals into a consistent proxy world by reasoning over all views, and then refines it through render-and-compare with numerical fitting.
\item A video renderer fine-tuned to follow the simulated proxy frames while taking the appearance of the scene from its context views, so that the same world supports both spatial exploration and physical interaction.
\end{itemize}

\begin{figure*}[t]
    \centering
    \includegraphics[width=\textwidth]{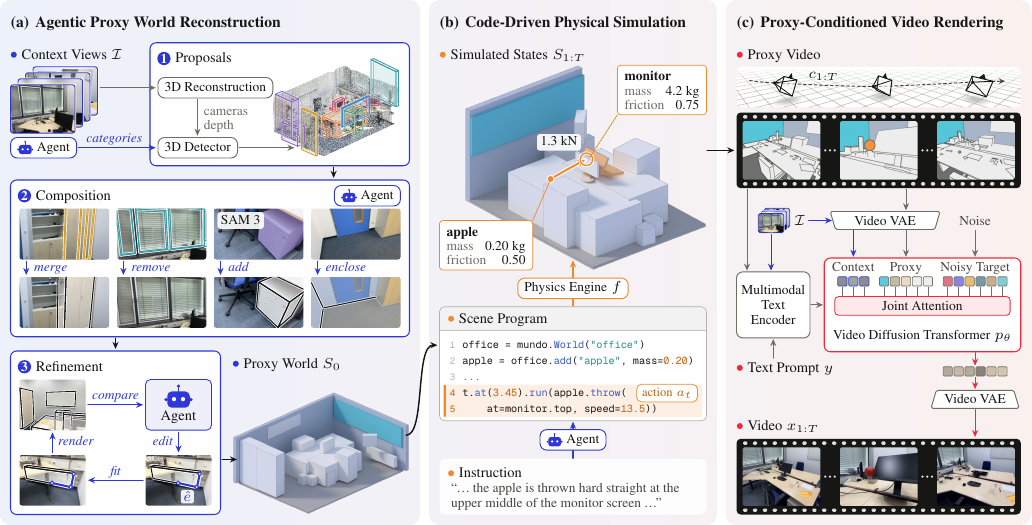}
    \caption{\textbf{\NICKNAME{} reconstructs a place as a proxy world, simulates it, and renders the result as a video.} (a) The agent composes the box proposals of a 3D detector into one scene and refines it into the proxy world $S_0$ by render-and-compare with numerical fitting. (b) Given an instruction, the agent writes the actions $a_t$ into the scene program, which a physics engine executes into the states $S_{1:T}$. (c) A video diffusion transformer renders the proxy video, rasterized along the camera trajectory $c_{1:T}$, into the video $x_{1:T}$.}
    \label{fig:pipeline}
\end{figure*}

\section{Related Work}
\label{sec:related}

\noindent\textbf{Video world models.}
Video world models generate observations from history and actions~\cite{bruce2024genie,alonso2024diamond,valevski2024gamengen,yu25gamefactory,parker-holder25genie,nvidia2025cosmos,robbyant2026lingbotworld,xiao2026worldmem}, building on large video generators~\cite{wan2025wan,minimax2026h3} that are becoming general visual learners~\cite{wiedemer25video}.
The generated scene drifts over long rollouts and when the camera returns~\cite{chen24diffusion,zhang2025framepack}.
Remedies change how history frames are used~\cite{chen24diffusion,song25history-guided,huang25self,zhang2025framepack}, retrieve past frames from memory~\cite{xiao2026worldmem,yu2025cam,li25vmem}, or condition on renderings of explicit 3D~\cite{wu25video,yu2024viewcrafter,ren2025gen3c,lan2026holo4d,liao2026puffin}.
However, the world remains inside the model and cannot be edited.
Instead, \NICKNAME{} keeps it in an executable program that the video model only renders.

\noindent\textbf{Physics-grounded video generation.}
Some video diffusion models learn motion from simulation data or condition on forces and motion~\cite{li2025pisa,wang2025physctrl,wang26streamforce,burgert2025gowiththeflow,aigcapps2026videoxfun}, but leave physical parameters implicit.
Simulators expose them, but direct rendering needs full appearance models~\cite{xie2024physgaussian,zhang2024physdreamer,chen2025physgen3d}.
Recent work renders simulated motion with a video generator~\cite{liu2024physgen,foo2026psivg,li2025wonderplay,zhan2026perpetualwonder,liu2026realwonder,mirros2026agentgarten}.
These scenes mostly come from a single image or template and cover only what it shows.
\NICKNAME{} reconstructs a whole place from several views, so one world supports both exploration and interaction.

\noindent\textbf{Feed-forward scene reconstruction.}
Feed-forward models recover geometry~\cite{wang2025vggt,wang2026pi3,lin2026da3,lan26stream3r,luo20264rc}, 3D Gaussians~\cite{charatan23pixelsplat,chen2024mvsplat,jiang2025anysplat}, object boxes~\cite{lazarow2025cubify,detone2026boxer,mao2025spatiallm}, or unseen views with diffusion priors~\cite{wu2023reconfusion,gao2024cat3d,zhou2025seva,wu2026framecrafter}.
These outputs give geometry and appearance, but no object identity, physical properties, or rules, so they cannot be simulated or edited.
A single pass can also miss, duplicate, or misplace objects.
When they drive a video model, motion and interaction still come from that model or from a source video~\cite{bai2025recammaster,yu2025trajectorycrafter,lin2026vista4d,lan2026holo4d}.

\noindent\textbf{Coding agents for scene reconstruction.}
Coding agents write, run, and revise programs, and check them against images~\cite{openai2026gpt6astra,anthropic2026claudeopus55}.
Code is explicit and editable, and can be checked by executing and rendering it.
Code world models let agents write worlds for video models to render, but invent them rather than recover a captured place~\cite{chen2026cwm,huang2026pwm,liang2026codevideomodel,fang2026coder}.
For reconstruction, agents refine shapes by render and compare~\cite{mazur2026agentstar} or compile scans into simulation-ready rooms~\cite{huang2026literealityagent}, and real-to-sim methods aim for detailed assets~\cite{chen2024urdformer,torne2024rialto,mandi2024real2code}.
An agent can compare renders with several views and fix errors across them, which one feed-forward pass cannot.
\NICKNAME{} therefore uses a coding agent to correct detector proposals~\cite{detone2026boxer,carion2026sam3} across views, keeping only the layout, collision geometry, and physical parameters that simulation needs.

\section{Method}
\label{sec:method}

As shown in \cref{fig:pipeline}, \NICKNAME{} connects three components through an explicit 3D proxy world.
We describe how an \emph{agent} reconstructs the world from the context views in \cref{sec:method_recon}, how a \emph{physics engine} advances it by executing a scene program in \cref{sec:method_code}, and how a \emph{video diffusion model} renders it into video in \cref{sec:method_render}.

\subsection{Overview}
\label{sec:method_overview}
A world model consists of a state, the actions that change it, a transition, and an observation.
In \NICKNAME{}, the \emph{state} $S_t$ is the proxy world: the objects of the scene with their categories, poses, sizes, and physical properties, together with the enclosing surfaces.
An \emph{action} $a_t$ is a change applied to this world, such as a force, an initial velocity, a scripted motion, or a new value of a global parameter such as gravity; it is given directly or derived from a text instruction.
The \emph{observation} is a video $x_{1:T}$ of the world, viewed along a target camera trajectory $c_{1:T}$.
Given $K$ context views $\mathcal{I}=\{I_k\}_{k=1}^{K}$ of a scene, the agentic reconstruction $\mathcal{R}$ first recovers the initial state $S_0$ and the camera $\bar{c}_k$ of every context view,
\begin{equation}
    \big(S_0,\ \{\bar{c}_k\}_{k=1}^{K}\big) = \mathcal{R}(\mathcal{I}).
    \label{eq:reconstruct}
\end{equation}
The state then evolves under the actions,
\begin{equation}
    S_{t+1} = f(S_t, a_t),
    \label{eq:transition}
\end{equation}
where a physics engine computes the transition $f$.
Finally, a video diffusion model renders the states into the video $x_{1:T}$.
It learns the conditional distribution
\begin{equation}
    p_{\bm\theta}\big(x_{1:T} \mid \pi(S_{1:T}, c_{1:T}),\ \mathcal{I},\ y\big),
    \label{eq:render}
\end{equation}
where $\pi$ rasterizes the states into proxy frames along the target trajectory, and $y$ is the text prompt.

\subsection{Agentic Proxy World Reconstruction}
\label{sec:method_recon}
\paragraph{Proxy representation.}
A world state does not need to encode fine appearance: JEPA world models predict in an abstract representation and discard pixel detail~\cite{lecun22a-path,assran25v-jepa}.
In \NICKNAME{}, this abstraction is explicit: the proxy records where each object is and how much space it occupies, not what it looks like.
Specifically, each object is represented by a semantic category and a 3D bounding box $b_i=(\mathbf{p}_i,\mathbf{s}_i,\psi_i)$ with center $\mathbf{p}_i\in\mathbb{R}^3$, size $\mathbf{s}_i\in\mathbb{R}^3_{+}$, and yaw $\psi_i$.
Planar surfaces represent the enclosing structure, such as the ground and walls.
Surface detail and appearance are left to the renderer, which observes them directly in the context views.

\paragraph{Reconstruction as an agentic process.}
Recovering this proxy from a few views is mostly a matter of deciding what the scene contains.
Which objects exist, which detections in different views show the same object, and where the room ends are questions about the whole scene, not about one image.
Recent vision-language models (VLMs)~\cite{openai2026gpt6astra,anthropic2026claudeopus55} have improved rapidly at this kind of spatial reasoning.
We therefore let a VLM agent build the proxy world in three steps, described next: (i)~it gathers box \emph{proposals}, (ii)~\emph{composes} them into a scene, and (iii)~\emph{refines} the scene by rendering and comparing.

\paragraph{(i) Proposals.}
\NICKNAME{} obtains the cameras $\{\bar{c}_k\}$ and dense depth of the context views either from the capture or from a 3D foundation model~\cite{wang2026pi3}.
The agent lists the object categories it sees in the views, and an off-the-shelf 3D detector~\cite{detone2026boxer}, prompted with these categories, proposes boxes.
The proposals are geometrically accurate but predicted in a single pass: they miss some objects, duplicate others across views, include spurious boxes, and usually lack the complete enclosing structure.
The detector does not enforce consistency across views or over the whole scene.

\paragraph{(ii) Composition.}
The agent composes the proposals into one scene by reasoning over all views at once.
It decides which boxes to keep, merge as duplicates of one object, or remove as spurious.
For objects missed by the detector, it adds boxes lifted from the masks of SAM~3~\cite{carion2026sam3}, prompted with the same categories: each mask is back-projected to 3D with the recovered depth, and a box is fitted to its points.
It also adds the enclosing surfaces, with the ground fitted as a plane to the recovered depth and the walls placed along the room boundary observed in the views.
The agent decides what the scene should contain, and the recovered depth determines where each object is placed.
A user can steer these decisions with an optional text prompt, \eg, to also include the paintings on a wall.

\paragraph{(iii) Render-and-compare refinement.}
The agent then checks the scene against the images using scene-editing and numerical fitting tools~\cite{mazur2026agentstar}.
In each iteration, we render the current proxy from the context cameras and overlay the renders on the views, allowing the agent to locate mismatches.
It edits the scene where objects are still missing, duplicated, or spurious.
For a box with an inaccurate pose or size, it calls the fitting tool, leaving metric estimation to a solver.
The agent marks visible object edges $\hat{e}$ in a set of views $\mathcal{V}_{\text{fit}}$ and selects the box parameters $\phi_i\subseteq\{\mathbf{p}_i,\mathbf{s}_i,\psi_i\}$ to update.
A least-squares solver then optimizes
\begin{equation}
    \phi_i^{\star} = \argmin_{\phi_i} \sum_{v\in\mathcal{V}_{\text{fit}}} \sum_{e} \rho\Big(d\big(\hat{e}_{v},\ \Pi_{\bar{c}_v}(e;\, b_i)\big)\Big),
    \label{eq:fit}
\end{equation}
where $\Pi_{\bar{c}_v}(e; b_i)$ projects edge $e$ of box $b_i$ into view $v$, $d$ is the image-space distance between the marked and projected edges, and $\rho$ is a robust loss.
A fit is accepted only if it also agrees with the remaining context views outside $\mathcal{V}_{\text{fit}}$.
The loop repeats until a separate reviewer agent, which inspects the overlays on all context views, accepts the proxy or a fixed budget of tool calls and time is exhausted.

\begin{figure*}[t]
    \centering
    \includegraphics[width=\textwidth]{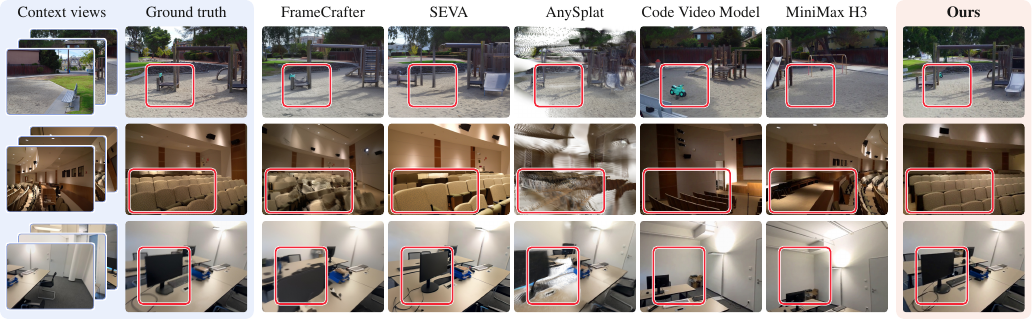}
    \caption{\textbf{Qualitative comparison of camera-controlled video generation.} Each row shows three context views and the target view. As outlined in red, the baselines morph or misplace scene content, whereas \NICKNAME{} stays faithful to the ground truth.}
    \label{fig:nvs}
    \vspace{-3mm}
\end{figure*}

\subsection{Code-Driven Physical Simulation}
\label{sec:method_code}
\NICKNAME{} stores the proxy world as an executable scene program.
The program declares every object with its box, category, and physical properties such as mass, friction, and restitution.
It also schedules the actions $a_t$ on a timeline: applied forces, initial velocities, scripted motions, and changes to global parameters such as gravity.
Given a text instruction, the agent writes these actions and sets the physical parameters they require.
Users can also edit the program to add an object, alter a property, or redirect the camera.
A physics engine~\cite{todorov2012mujoco} executes the program to compute $f$ in \cref{eq:transition}, so the motion and contact in the states $S_{1:T}$ are physically grounded.
In spatial exploration, only the camera moves: the state stays static and the box proxy suffices.
In physical interaction, objects in contact are upgraded from boxes to finer geometry, which the agent composes from primitives, obtains from a 3D generator~\cite{xiang2025trellis,chen2026sam3d}, or imports as existing simulation-ready assets.

\subsection{Proxy-Conditioned Video Rendering}
\label{sec:method_render}
\paragraph{Proxy video.}
The states reach the video model as a proxy video~\cite{chen2026cwm}: for every frame, $\pi$ rasterizes the state $S_t$ from the target camera $c_t$ into a proxy frame.
Every object is drawn in a flat, unshaded color indexed by its semantic category.
The proxy video thus fixes the layout, motion, and camera trajectory of every frame, while the context views provide the appearance.
The appendix gives more details of the proxy rendering.

\paragraph{Renderer architecture.}
We fine-tune a pretrained video diffusion transformer~\cite{minimax2026h3} as the renderer.
The video VAE encodes the proxy video and the context views, and their tokens are appended to the noisy target tokens, so every transformer block attends jointly over them.
The context views and subsampled proxy frames are also fed to the multimodal text encoder together with the text prompt $y$, adding semantic guidance to the pixel-level VAE tokens.
Since the context views cannot show newly inserted objects (\textit{e.g.}, a robot), we provide reference images of these objects in the same way as the context views.

\section{Experiments}
\label{sec:exp}

We first describe the implementation details.
We then compare \NICKNAME{} with existing methods on proxy world reconstruction, camera-controlled video generation, and physics-grounded video generation, and ablate its key designs.

\paragraph{Implementation details.}
The agent is GPT-6 Astra~\cite{openai2026gpt6astra} with the Codex harness unless otherwise stated, and the physics engine is MuJoCo~\cite{todorov2012mujoco}.
The renderer fine-tunes MiniMax H3 Ref2VA~\cite{minimax2026h3} with rank-128 LoRA~\cite{hu2022lora} for 3K iterations on 56 NVIDIA A100 GPUs, using 124-frame videos at $640\times480$ with up to 12 context views.
Its training set contains about 40K pairs from eight real and synthetic sources~\cite{baruch2021arkitscenes,pan2023adt,chang2019argoverse,banerjee2025hot3d,chao2021dexycb,ahmadyan2021objectron,chen2026cwm,deitke2022procthor}, which are used exclusively for renderer training and are not involved in agentic reconstruction.
The data curation and more details are provided in the appendix.

\subsection{Proxy Reconstruction Evaluation}
\label{sec:exp_recon}
\begin{table}[t]
    \centering
    \caption{\textbf{3D proxy world reconstruction.} Scene-level, class-agnostic F1 (\%), with the mean and standard deviation of \NICKNAME{} over three runs. The \legendbest{best} and \legendsecond{second-best} results are highlighted.}
    \label{tab:reconstruction}
    \footnotesize
    \setlength{\tabcolsep}{2pt}
    \renewcommand{\arraystretch}{1.1}
    \begin{tabular*}{\linewidth}{@{\extracolsep{\fill}}lcccc}
        \toprule
        & \multicolumn{2}{c}{CA-1M~\cite{lazarow2025cubify}} & \multicolumn{2}{c}{ADT~\cite{pan2023adt}} \\
        \cmidrule(lr){2-3} \cmidrule(l){4-5}
        Method & F1@25 $\uparrow$ & F1@50 $\uparrow$ & F1@25 $\uparrow$ & F1@50 $\uparrow$ \\
        \midrule
        SpatialLM~\cite{mao2025spatiallm} & 7.1 & 2.9 & 10.9 & 7.9 \\
        CuTR~\cite{lazarow2025cubify} & 32.3 & 19.1 & 27.3 & 12.6 \\
        Boxer~\cite{detone2026boxer} & \tablesecond{41.0} & \tablesecond{21.2} & \tablesecond{35.5} & \tablesecond{19.7} \\
        \midrule
        \textbf{\NICKNAME{} (Ours)} & \tablebest{51.3\tablestd{0.6}} & \tablebest{27.1\tablestd{0.7}} & \tablebest{39.3\tablestd{0.6}} & \tablebest{20.9\tablestd{0.8}} \\
        \bottomrule
    \end{tabular*}
    \vspace{-3mm}
\end{table}

\begin{table*}[t]
    \centering
    \caption{\textbf{Camera-controlled video generation on ScanNet++ and Tanks and Temples.}}
    \label{tab:nvs}
    \small
    \setlength{\tabcolsep}{7pt}
    \renewcommand{\arraystretch}{1.0}
    \begin{tabular}{lcccccccc}
        \toprule
        & \multicolumn{4}{c}{ScanNet++~\cite{yeshwanth2023scannetpp}} & \multicolumn{4}{c}{Tanks and Temples~\cite{knapitsch2017tanks}} \\
        \cmidrule(lr){2-5} \cmidrule(l){6-9}
        Method & PSNR $\uparrow$ & SSIM $\uparrow$ & LPIPS $\downarrow$ & FID $\downarrow$
               & PSNR $\uparrow$ & SSIM $\uparrow$ & LPIPS $\downarrow$ & FID $\downarrow$ \\
        \midrule
        FramePack~\cite{zhang2025framepack} & 9.11 & 0.426 & 0.669 & 141.9 & 10.65 & 0.234 & 0.645 & 131.5 \\
        MiniMax H3~\cite{minimax2026h3} & 9.64 & 0.445 & 0.655 & 113.5 & 11.58 & 0.265 & 0.632 & 96.5 \\
        Code Video Model~\cite{liang2026codevideomodel} & 9.74 & 0.425 & 0.656 & 108.4 & 11.68 & 0.267 & 0.625 & 89.1 \\
        AnySplat~\cite{jiang2025anysplat} & 11.59 & 0.480 & 0.586 & 190.4 & 11.17 & 0.266 & 0.595 & 154.5 \\
        TrajectoryCrafter~\cite{yu2025trajectorycrafter} & 10.63 & 0.486 & 0.649 & 159.9 & 12.33 & 0.323 & 0.625 & 139.0 \\
        Holo4D~\cite{lan2026holo4d} & 12.04 & 0.543 & 0.551 & 136.2 & 13.74 & 0.346 & 0.541 & 92.0 \\
        SEVA~\cite{zhou2025seva} & \tablesecond{15.15} & \tablesecond{0.606} & \tablesecond{0.418} & \tablesecond{72.8}
                              & 13.94 & 0.356 & 0.455 & 58.3 \\
        Gen3C~\cite{ren2025gen3c} & 14.74 & 0.603 & 0.486 & 124.4 & 14.51 & \tablesecond{0.383} & 0.481 & 89.4 \\
        FrameCrafter~\cite{wu2026framecrafter} & 14.35 & 0.587 & 0.459 & 82.4
                              & \tablesecond{14.91} & 0.375 & \tablebest{0.450} & \tablesecond{57.8} \\
        \midrule
        \textbf{\NICKNAME{} (Ours)} & \tablebest{15.91} & \tablebest{0.622} & \tablebest{0.404} & \tablebest{63.3}
                       & \tablebest{15.41} & \tablebest{0.384} & \tablesecond{0.451} & \tablebest{54.0} \\
        \bottomrule
    \end{tabular}
    \vspace{-3mm}
\end{table*}

\paragraph{Setup.}
We evaluate the spatial and structural fidelity of the reconstructed 3D bounding boxes.
On 10 scenes of CA-1M~\cite{lazarow2025cubify} and 6 sequences of ADT~\cite{pan2023adt}, we follow Boxer~\cite{detone2026boxer} and evaluate scene-level 3D detections class-agnostically. We report F1 at 3D IoU thresholds of 0.25 and 0.5.
We compare with SpatialLM~\cite{mao2025spatiallm}, CuTR~\cite{lazarow2025cubify}, and Boxer~\cite{detone2026boxer}.
All methods are evaluated on the same frames with the ground-truth camera poses and depth maps provided by the datasets.
Since the agent is stochastic, we report the mean and standard deviation of \NICKNAME{} over three independent runs.

\paragraph{Results.}
As shown in \cref{tab:reconstruction}, \NICKNAME{} outperforms all baselines on both datasets and at both IoU thresholds.
On CA-1M, it surpasses the strongest baseline, Boxer, by 10.3 F1 at IoU 0.25 and by 5.9 F1 at IoU 0.5, a relative improvement of 25\% and 28\%, respectively.
The advantage carries over to ADT, where \NICKNAME{} reaches 39.3 and 20.9 F1 against 35.5 and 19.7 for Boxer.
The gain mainly comes from the visual reasoning of the agent, which merges duplicate proposals and removes spurious ones, as we analyze in \cref{sec:exp_ablation}.
The reconstructed boxes thus provide a reliable object-level layout for the proxy world.

\subsection{Spatial Exploration Evaluation}
\label{sec:exp_nvs}
\paragraph{Setup.}
We evaluate spatial exploration as camera-controlled video generation, \ie, rendering novel views of a captured scene along a target camera trajectory from a few posed context views.
Our test set contains 20 scenes from ScanNet++~\cite{yeshwanth2023scannetpp} and Tanks and Temples (T\&T)~\cite{knapitsch2017tanks}, with 10 test sequences on each dataset, covering both indoor and outdoor environments.
The number of context views is 9 on ScanNet++ and 8 on T\&T.
\NICKNAME{} reconstructs each proxy world from the context views with depth estimated by $\pi^3$~\cite{wang2026pi3}.
Following prior work~\cite{zhou2025seva,ren2025gen3c,yu2025cam}, we measure the consistency with the captured scene by PSNR, SSIM, and LPIPS~\cite{zhang2018lpips} between the generated and the ground-truth frames, and the visual quality by FID~\cite{heusel2017fid}.
We compare with recent camera-controlled generative methods, including Stable Virtual Camera (SEVA)~\cite{zhou2025seva}, Gen3C~\cite{ren2025gen3c}, FrameCrafter~\cite{wu2026framecrafter}, Holo4D~\cite{lan2026holo4d}, and TrajectoryCrafter~\cite{yu2025trajectorycrafter}.
We also include the feed-forward 3D reconstruction method AnySplat~\cite{jiang2025anysplat}, as well as FramePack~\cite{zhang2025framepack} and the pretrained MiniMax H3~\cite{minimax2026h3}, which take the context views as visual references.
We further compare with Code Video Model~\cite{liang2026codevideomodel}, a training-free method that guides the frozen MiniMax H3 with the same proxy videos as \NICKNAME.
Each baseline follows its native input setting, with all inputs derived from the same context views and target camera trajectories.
Notably, Holo4D operates on a source video, and we provide it with a captured video clip of the scene, which is denser than our context views.

\paragraph{Results.}
Quantitative results are reported in \cref{tab:nvs}.
\NICKNAME{} achieves the best results on almost all metrics across the two datasets.
This demonstrates that its videos are both consistent with the captured scene and visually realistic.
Code Video Model guides the frozen base model with the same proxy videos, yet its results neither follow the target camera trajectory nor stay consistent with the context views.
The advantage of \NICKNAME{} is even more evident in the qualitative comparisons of \cref{fig:nvs}.
FrameCrafter and SEVA, two of the strongest baselines, are roughly consistent with the scene in overall structure, but they morph objects or misplace scene content, \eg, the seats of the auditorium.
In contrast, \NICKNAME{} follows the target viewpoint and faithfully preserves the scene content, thanks to the proxy frames that specify the viewpoint and layout of every target frame.

\subsection{Physical Interaction Evaluation}
\label{sec:exp_physics}
\begin{table}[t]
    \centering
    \caption{\textbf{Physics-grounded video generation.} Trajectory IoU and error, the latter in pixels with all videos resized to $512\times384$, measure the fidelity to the simulation. A VLM judge rates physical realism, photorealism, and semantic consistency in $[0,1]$.}
    \label{tab:physics}
    \footnotesize
    \setlength{\tabcolsep}{1.0pt}
    \renewcommand{\arraystretch}{1.1}
    \begin{tabular*}{\linewidth}{@{\extracolsep{\fill}}l@{\hspace{3pt}}lccccc}
        \toprule
        & & \multicolumn{2}{c}{Trajectory} & \multicolumn{3}{c}{VLM judge} \\
        \cmidrule(lr){3-4} \cmidrule(l){5-7}
        Method & Condition & IoU $\uparrow$ & Err. $\downarrow$ & Phys. $\uparrow$ & Photo. $\uparrow$ & Sem. $\uparrow$ \\
        \midrule
        MiniMax H3~\cite{minimax2026h3} & text & 0.29 & 93.9 & 0.39 & \tablebest{0.91} & 0.44 \\
        Wan2.2 I2V-A14B~\cite{wan2025wan} & text & 0.21 & 107.0 & 0.49 & 0.83 & 0.56 \\
        PhysGen3D~\cite{chen2025physgen3d} & params. & 0.32 & 82.9 & 0.50 & 0.75 & 0.58 \\
        Go-with-the-Flow~\cite{burgert2025gowiththeflow} & sim.\ flow & 0.51 & 34.8 & 0.66 & 0.77 & 0.75 \\
        Wan2.2-Fun A14B~\cite{aigcapps2026videoxfun} & sim.\ depth & \tablesecond{0.78} & \tablesecond{19.3} & 0.66 & 0.84 & 0.75 \\
        Code Video Model~\cite{liang2026codevideomodel} & sim.\ proxy & 0.57 & 31.0 & \tablesecond{0.70} & \tablesecond{0.90} & \tablesecond{0.77} \\
        \midrule
        \textbf{\NICKNAME{} (Ours)} & sim.\ proxy & \tablebest{0.83} & \tablebest{12.7} & \tablebest{0.75} & \tablebest{0.91} & \tablebest{0.81} \\
        \bottomrule
    \end{tabular*}
    \vspace{-3mm}
\end{table}

\begin{figure*}[t]
    \centering
    \includegraphics[width=\textwidth]{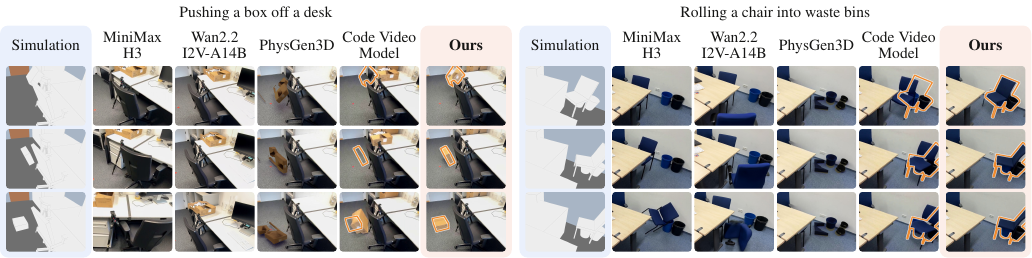}
    \caption{\textbf{Qualitative comparison of physics-grounded video generation.} The first column of each interaction shows the simulation that specifies it, and time runs downward. The text-conditioned models do not follow the specified physics, and PhysGen3D distorts the objects. With the simulated objects outlined in orange, Code Video Model drifts away from them, whereas \NICKNAME{} follows them.}
    \label{fig:physics}
    \vspace{-3mm}
\end{figure*}

\paragraph{Setup.}
Following the evaluation protocol of PhysGen3D~\cite{chen2025physgen3d}, we build a benchmark of 20 interaction scenarios on real and generated scenes, covering dropping, sliding, collision, throwing, and altered gravity.
Each scenario specifies the moving objects with their initial velocities and the physical parameters, \ie, friction, restitution, and gravity.
Such a specification determines the motion, also under conditions that cannot be captured, such as altered gravity, so the reference of each scenario is its outcome under rigid-body dynamics, computed by a physics engine~\cite{todorov2012mujoco}.
We then evaluate two complementary aspects: whether a video follows the specified interaction, and whether its motion is physically realistic when judged independently of our simulation.
For the former, we track each moving object in the generated video with SAM~2~\cite{ravi2025sam2} and report its bounding-box IoU with the simulated object as trajectory IoU and the distance between their displacements as trajectory error.
For the latter, a VLM judge~\cite{bai2025qwen3vl}, which has no access to the simulation, rates physical realism, \ie, whether the motion obeys real-world physics, together with photorealism and semantic consistency as in PhysGen3D~\cite{chen2025physgen3d}, with details in the appendix.
We compare with three families of methods: the text-conditioned video generation models MiniMax H3~\cite{minimax2026h3} and Wan2.2 I2V-A14B~\cite{wan2025wan}, which receive the interaction as a text prompt, the simulation-based PhysGen3D~\cite{chen2025physgen3d}, and the controllable video generation models Wan2.2-Fun A14B~\cite{aigcapps2026videoxfun}, Go-with-the-Flow~\cite{burgert2025gowiththeflow}, and Code Video Model~\cite{liang2026codevideomodel}.
The controllable models are driven by the same simulation as \NICKNAME, which isolates how faithfully each model renders the simulated motion.
The input of each method is listed in \cref{tab:physics}.

\paragraph{Results.}
As shown in \cref{tab:physics}, \NICKNAME{} follows the simulation most faithfully, with the highest trajectory IoU of 0.83 and the lowest trajectory error of 12.7 pixels, and achieves the best physical realism and semantic consistency, while matching the photorealism score of MiniMax H3 at 0.91.
Text-conditioned models such as H3 produce visually appealing videos but fail to follow the specified physical conditions, with a trajectory IoU below 0.3 and a trajectory error above 90 pixels.
For example, in \cref{fig:physics}, the pushed box is still on the desk when it should already lie on the floor, and the chair topples over instead of rolling into the bins.
PhysGen3D has to generate the 3D shape of each object from a single view, so the unobserved sides are hallucinated and the object appears distorted once it moves.
Controllable models benefit directly from our proxy world: driven by our simulation, they reach a trajectory IoU of 0.51 to 0.78 and become the strongest baselines in physical realism.
However, none of them follows the simulation as faithfully as \NICKNAME.
Notably, Code Video Model receives the same proxy video as \NICKNAME, yet its trajectory IoU drops to 0.57 and its objects drift away from the simulated motion.
In contrast, the videos generated by \NICKNAME{} faithfully obey the simulated physics.

\subsection{Ablation Studies}
\label{sec:exp_ablation}

\begin{table}[t]
    \centering
    \caption{\textbf{Ablation of proxy world reconstruction on CA-1M.} Top: the steps of our agentic reconstruction, added in turn to Boxer~\cite{detone2026boxer}, in gray. Below: other agent models and input geometry. Mean and standard deviation over three runs.}
    \label{tab:reconstruction_ablation}
    \small
    \setlength{\tabcolsep}{4pt}
    \begin{tabular}{@{}p{0.535\linewidth}@{}>{\centering\arraybackslash}p{0.2325\linewidth}@{}>{\centering\arraybackslash}p{0.2325\linewidth}@{}}
        \toprule
        Variant & F1@25 $\uparrow$ & F1@50 $\uparrow$ \\
        \midrule
        \multicolumn{3}{@{}l}{\textit{Agentic reconstruction}} \\
        \rowcolor{tablerefgray}[0pt][0pt]\hspace{0.8em}Boxer~\cite{detone2026boxer} & 41.0 & 21.2 \\
        \hspace{0.8em}+ Agent-identified categories & 43.6\tablestd{0.3} & 24.3\tablestd{0.7} \\
        \hspace{0.8em}+ Composition & 50.0\tablestd{0.5} & 26.9\tablestd{0.6} \\
        \hspace{0.8em}\textbf{+ Refinement (Ours)} & 51.3\tablestd{0.6} & 27.1\tablestd{0.7} \\
        \midrule
        \multicolumn{3}{@{}l}{\textit{Agent model}} \\
        \hspace{0.8em}w/ GPT-5.6 Sol & 43.5\tablestd{0.6} & 21.6\tablestd{0.4} \\
        \hspace{0.8em}w/ Claude Opus 5.5 & 50.5\tablestd{0.4} & 27.5\tablestd{0.7} \\
        \midrule
        \multicolumn{3}{@{}l}{\textit{Input geometry}} \\
        \hspace{0.8em}w/ $\pi^3$ cameras and depth & 49.8\tablestd{0.1} & 23.9\tablestd{0.6} \\
        \bottomrule
    \end{tabular}
    \vspace{-3mm}
\end{table}

\paragraph{Agentic reconstruction.}
We first examine the contribution of the agent to the reconstruction on CA-1M in \cref{tab:reconstruction_ablation}.
Taking Boxer~\cite{detone2026boxer} as the reference, prompting the detector with the object categories that the agent identifies in the scene already improves F1 at IoU 0.25 from 41.0 to 43.6.
Composing the proxy world with the agent then brings the largest gain, from 43.6 to 50.0.
In this step, the agent identifies the objects in the context views, complements the proposals with SAM~3, and decides through visual reasoning which boxes to keep, merge, or remove.
The render-and-compare refinement with numerical fitting further improves F1 at IoU 0.25 to 51.3.
These results verify that the visual reasoning of the agent is the key to an accurate proxy world.

\paragraph{Agent model.}
We further replace the agent model while keeping the proposals and the tools unchanged.
As shown in \cref{tab:reconstruction_ablation}, Claude Opus 5.5 performs comparably to our default GPT-6 Astra, which shows that \NICKNAME{} is not tied to a particular agent model.
Moreover, moving from GPT-5.6 Sol~\cite{openai2026gpt56} to the newer GPT-6 Astra improves F1 at IoU 0.25 from 43.5 to 51.3.
The reconstruction could thus further improve as agent models advance.

\paragraph{Input geometry.}
The results above take the cameras and depth provided by the dataset.
When both are instead estimated by $\pi^3$~\cite{wang2026pi3} from the RGB frames alone, \NICKNAME{} still reaches 49.8 F1 at IoU 0.25, close to the 51.3 with ground-truth geometry and above all baselines on CA-1M, which take ground-truth depth.

\begin{table}[t]
    \centering
    \caption{\textbf{Ablation of proxy-conditioned rendering on Tanks and Temples.} Top: two variants of the renderer fine-tuned from the same initialization, without and with the proxy video. Bottom: \NICKNAME{} with different numbers of context views.}
    \label{tab:rendering_ablation}
    \small
    \setlength{\tabcolsep}{4pt}
    \begin{tabular*}{\linewidth}{@{\extracolsep{\fill}}lcccc}
        \toprule
        Variant & PSNR $\uparrow$ & SSIM $\uparrow$ & LPIPS $\downarrow$ & FID $\downarrow$ \\
        \midrule
        \multicolumn{5}{@{}l}{\textit{Proxy video}} \\
        \hspace{0.8em}w/o proxy video & 11.64 & 0.269 & 0.630 & 101.1 \\
        \hspace{0.8em}w/ proxy video & 15.14 & 0.374 & 0.458 & 54.4 \\
        \midrule
        \multicolumn{5}{@{}l}{\textit{Number of context views}} \\
        \hspace{0.8em}1 & 13.73 & 0.347 & 0.512 & 90.4 \\
        \hspace{0.8em}3 & 15.11 & 0.380 & 0.458 & 58.9 \\
        \hspace{0.8em}8 & 15.41 & 0.384 & 0.451 & 54.0 \\
        \bottomrule
    \end{tabular*}
    \vspace{-3mm}
\end{table}

\begin{figure}[t]
    \centering
    \includegraphics[width=\linewidth]{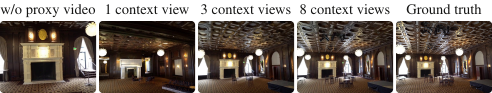}
    \caption{\textbf{Qualitative ablations of rendering.} The proxy video anchors the rendered view to the target viewpoint, and the rendered content becomes progressively closer to the ground truth as more context views are provided.}
    \label{fig:ablation}
    \vspace{-1mm}
\end{figure}

\paragraph{Proxy video.}
To validate the proxy video, we fine-tune two variants of the renderer from the same initialization for 3K iterations on 16 A100 GPUs, with and without the proxy video, where the latter instead injects the cameras as embeddings into the context-view and video latents.
All other settings are kept the same.
As shown in \cref{tab:rendering_ablation}, removing the proxy video degrades all metrics by a large margin, \eg, PSNR drops from 15.14\,dB to 11.64\,dB.
The proxy video provides the geometric layout of the scene in every target frame, which helps the renderer localize the target view and place the scene content accordingly, as shown in \cref{fig:ablation}.

\paragraph{Number of context views.}
We also ablate the number of context views provided to the renderer.
As shown in \cref{tab:rendering_ablation} and \cref{fig:ablation}, the fidelity improves consistently with more context views, as they provide more evidence of the scene.

\begin{figure}[t]
    \centering
    \includegraphics[width=\linewidth]{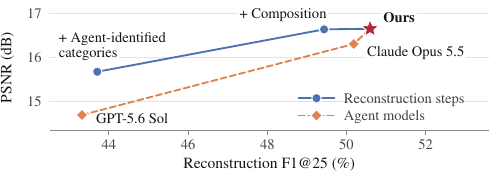}
    \caption{\textbf{Rendering quality improves with reconstruction quality.} Each point renders the CA-1M scenes from the reconstruction of one variant in \cref{tab:reconstruction_ablation} with the same renderer.}
    \label{fig:recon_render}
    \vspace{-1mm}
\end{figure}

\paragraph{Reconstruction quality.}
Finally, we study how the quality of the proxy world affects rendering.
We render the CA-1M test scenes, which are excluded from the training data of the renderer, from the reconstructions of the variants in \cref{tab:reconstruction_ablation} with the same renderer, and plot the rendering PSNR against the reconstruction F1 in \cref{fig:recon_render}.
The rendering quality improves with the reconstruction quality, both along the steps of our agentic reconstruction and across agent models, as a more accurate proxy world provides a more faithful layout for rendering.
For example, replacing GPT-5.6 Sol with GPT-6 Astra as the agent improves PSNR from 14.70\,dB to 16.65\,dB.
\NICKNAME{} could therefore directly benefit from future advances in agent models and 3D reconstruction.

\section{Conclusion}
\label{sec:conclusion}

We introduce \NICKNAME, a customizable world model that turns context views of a scene into a world to explore and physically interact with.
Its world state is explicit: an agent recovers an executable scene program, a physics engine executes it, and a video model renders the result.
\NICKNAME{} outperforms existing methods in proxy world reconstruction and camera-controlled video generation, and its videos follow the specified physics.
Our results suggest that a world model of a specific scene need not be learned end to end: its components communicate only through the scene program and the proxy frames, so \NICKNAME{} can directly inherit advances in vision-language models, 3D foundation models, and video foundation models.
By leaving physics to simulation and appearance to generation, \NICKNAME{} makes world models controllable without sacrificing realism, offering an explicit, executable paradigm for the next generation of world models.

{
    \small
    \bibliographystyle{ieeenat_fullname}
    \bibliography{main,experiments}

\begin{thebibliography}{97}
\providecommand{\natexlab}[1]{#1}
\providecommand{\url}[1]{\texttt{#1}}
\expandafter\ifx\csname urlstyle\endcsname\relax
  \providecommand{\doi}[1]{doi: #1}\else
  \providecommand{\doi}{doi: \begingroup \urlstyle{rm}\Url}\fi

\bibitem[Ahmadyan et~al.(2021)Ahmadyan, Zhang, Ablavatski, Wei, and
  Grundmann]{ahmadyan2021objectron}
Adel Ahmadyan, Liangkai Zhang, Artsiom Ablavatski, Jianing Wei, and Matthias
  Grundmann.
\newblock Objectron: A large scale dataset of object-centric videos in the wild
  with pose annotations.
\newblock In \emph{Proc. {CVPR}}, 2021.

\bibitem[{Alibaba PAI}(2026)]{aigcapps2026videoxfun}
{Alibaba PAI}.
\newblock {VideoX-Fun}: A video generation pipeline for diffusion transformer.
\newblock \url{https://github.com/aigc-apps/VideoX-Fun}, 2026.

\bibitem[Alonso et~al.(2024)Alonso, Jelley, Micheli, Kanervisto, Storkey,
  Pearce, and Fleuret]{alonso2024diamond}
Eloi Alonso, Adam Jelley, Vincent Micheli, Anssi Kanervisto, Amos Storkey, Tim
  Pearce, and Fran{\c{c}}ois Fleuret.
\newblock Diffusion for world modeling: Visual details matter in {Atari}.
\newblock In \emph{Proc. {NeurIPS}}, 2024.

\bibitem[{Anthropic}(2026)]{anthropic2026claudeopus55}
{Anthropic}.
\newblock System card: {Claude Opus 5.5}.
\newblock \url{https://anthropic.com/claude-opus-5-5-system-card}, 2026.

\bibitem[Assran et~al.(2025)Assran, Bardes, Fan, Garrido, Howes, Komeili,
  Muckley, Rizvi, Roberts, Sinha, Zholus, Arnaud, Gejji, Martin, Hogan, Dugas,
  Bojanowski, Khalidov, Labatut, Massa, Szafraniec, Krishnakumar, Li, Ma,
  Chandar, Meier, LeCun, Rabbat, and Ballas]{assran25v-jepa}
Mido Assran, Adrien Bardes, David Fan, Quentin Garrido, Russell Howes, Mojtaba
  Komeili, Matthew~J. Muckley, Ammar Rizvi, Claire Roberts, Koustuv Sinha,
  Artem Zholus, Sergio Arnaud, Abha Gejji, Ada Martin, Francois~Robert Hogan,
  Daniel Dugas, Piotr Bojanowski, Vasil Khalidov, Patrick Labatut, Francisco
  Massa, Marc Szafraniec, Kapil Krishnakumar, Yong Li, Xiaodong Ma, Sarath
  Chandar, Franziska Meier, Yann LeCun, Michael Rabbat, and Nicolas Ballas.
\newblock V-{JEPA} 2: Self-supervised video models enable understanding,
  prediction and planning.
\newblock \emph{arXiv}, 2506.09985, 2025.

\bibitem[Bai et~al.(2025{\natexlab{a}})Bai, Xia, Fu, Wang, Mu, Cao, Liu, Hu,
  Bai, Wan, and Zhang]{bai2025recammaster}
Jianhong Bai, Menghan Xia, Xiao Fu, Xintao Wang, Lianrui Mu, Jinwen Cao, Zuozhu
  Liu, Haoji Hu, Xiang Bai, Pengfei Wan, and Di Zhang.
\newblock {ReCamMaster}: Camera-controlled generative rendering from a single
  video.
\newblock In \emph{Proc. {ICCV}}, 2025{\natexlab{a}}.

\bibitem[Bai et~al.(2025{\natexlab{b}})Bai, Cai, Chen, Chen, Chen, Cheng, Deng,
  Ding, Gao, Ge, Ge, Guo, Huang, Huang, Huang, Hui, Jiang, Li, Li, Li, Li, Lin,
  Lin, Liu, Liu, Liu, Liu, Liu, Liu, Lu, Luo, Lv, Men, Meng, Ren, Ren, Song,
  Sun, Tang, Tu, Wan, Wang, Wang, Wang, Wang, Xie, Xu, Xu, Xu, Yang, Yang,
  Yang, Yang, Yu, Zhang, Zhang, Zhang, Zheng, Zhong, Zhou, Zhou, Zhou, Zhu, and
  Zhu]{bai2025qwen3vl}
Shuai Bai, Yuxuan Cai, Ruizhe Chen, Keqin Chen, Xionghui Chen, Zesen Cheng,
  Lianghao Deng, Wei Ding, Chang Gao, Chunjiang Ge, Wenbin Ge, Zhifang Guo,
  Qidong Huang, Jie Huang, Fei Huang, Binyuan Hui, Shutong Jiang, Zhaohai Li,
  Mingsheng Li, Mei Li, Kaixin Li, Zicheng Lin, Junyang Lin, Xuejing Liu,
  Jiawei Liu, Chenglong Liu, Yang Liu, Dayiheng Liu, Shixuan Liu, Dunjie Lu,
  Ruilin Luo, Chenxu Lv, Rui Men, Lingchen Meng, Xuancheng Ren, Xingzhang Ren,
  Sibo Song, Yuchong Sun, Jun Tang, Jianhong Tu, Jianqiang Wan, Peng Wang,
  Pengfei Wang, Qiuyue Wang, Yuxuan Wang, Tianbao Xie, Yiheng Xu, Haiyang Xu,
  Jin Xu, Zhibo Yang, Mingkun Yang, Jianxin Yang, An Yang, Bowen Yu, Fei Zhang,
  Hang Zhang, Xi Zhang, Bo Zheng, Humen Zhong, Jingren Zhou, Fan Zhou, Jing
  Zhou, Yuanzhi Zhu, and Ke Zhu.
\newblock {Qwen3-VL} technical report.
\newblock \emph{arXiv}, 2511.21631, 2025{\natexlab{b}}.

\bibitem[Ball et~al.(2025)Ball, Bauer, Belletti, Brownfield, Ephrat, Fruchter,
  Gupta, Holsheimer, Ho{\l}y{\'n}ski, Hron, Kaplanis, Limont, McGill, Oliveira,
  Parker-Holder, Perbet, Scully, Shar, Spencer, Tov, Villegas, Wang, Yung,
  Baetu, Berbel, Bridson, Bruce, Buttimore, Chakera, Chandra, Collins, Cullum,
  Damoc, Dasagi, Gazeau, Gbadamosi, Han, Hirst, Kachra, Kerley, Kjems,
  Knoepfel, Koriakin, Lo, Lu, Mehring, Moufarek, Nandwani, Oliveira, Pardo,
  Park, Pierson, Poole, Ran, Salimans, Sanchez, Saprykin, Shen, Sidhwani,
  Smith, Stanton, Tomlinson, Vijaykumar, Wang, Wingfield, Wong, Xu, Yew, Young,
  Zubov, Eck, Erhan, Kavukcuoglu, Hassabis, Ghahramani, Hadsell, van~den Oord,
  Mosseri, Bolton, Singh, and Rockt{\"a}schel]{parker-holder25genie}
Philip~J. Ball, Jakob Bauer, Frank Belletti, Bethanie Brownfield, Ariel Ephrat,
  Shlomi Fruchter, Agrim Gupta, Kristian Holsheimer, Aleksander
  Ho{\l}y{\'n}ski, Jiri Hron, Christos Kaplanis, Marjorie Limont, Matt McGill,
  Yanko Oliveira, Jack Parker-Holder, Frank Perbet, Guy Scully, Jeremy Shar,
  Stephen Spencer, Omer Tov, Ruben Villegas, Emma Wang, Jessica Yung, Cip
  Baetu, Jordi Berbel, David Bridson, Jake Bruce, Gavin Buttimore, Sarah
  Chakera, Bilva Chandra, Paul Collins, Alex Cullum, Bogdan Damoc, Vibha
  Dasagi, Maxime Gazeau, Charles Gbadamosi, Woohyun Han, Ed Hirst, Ashyana
  Kachra, Lucie Kerley, Kristian Kjems, Eva Knoepfel, Vika Koriakin, Jessica
  Lo, Cong Lu, Zeb Mehring, Alex Moufarek, Henna Nandwani, Valeria Oliveira,
  Fabio Pardo, Jane Park, Andrew Pierson, Ben Poole, Helen Ran, Tim Salimans,
  Manuel Sanchez, Igor Saprykin, Amy Shen, Sailesh Sidhwani, Duncan Smith, Joe
  Stanton, Hamish Tomlinson, Dimple Vijaykumar, Luyu Wang, Piers Wingfield, Nat
  Wong, Keyang Xu, Christopher Yew, Nick Young, Vadim Zubov, Douglas Eck,
  Dumitru Erhan, Koray Kavukcuoglu, Demis Hassabis, Zoubin Ghahramani, Raia
  Hadsell, A{\"a}ron van~den Oord, Inbar Mosseri, Adrian Bolton, Satinder
  Singh, and Tim Rockt{\"a}schel.
\newblock {Genie 3}: A new frontier for world models.
\newblock
  \url{https://deepmind.google/blog/genie-3-a-new-frontier-for-world-models/},
  2025.

\bibitem[Banerjee et~al.(2025)Banerjee, Shkodrani, Moulon, Hampali, Han, Zhang,
  Zhang, Fountain, Miller, Basol, Newcombe, Wang, Engel, and
  Hodan]{banerjee2025hot3d}
Prithviraj Banerjee, Sindi Shkodrani, Pierre Moulon, Shreyas Hampali, Shangchen
  Han, Fan Zhang, Linguang Zhang, Jade Fountain, Edward Miller, Selen Basol,
  Richard Newcombe, Robert Wang, Jakob~Julian Engel, and Tomas Hodan.
\newblock {HOT3D}: Hand and object tracking in {3D} from egocentric multi-view
  videos.
\newblock In \emph{Proc. {CVPR}}, 2025.

\bibitem[Baruch et~al.(2021)Baruch, Chen, Dehghan, Dimry, Feigin, Fu, Gebauer,
  Joffe, Kurz, Schwartz, and Shulman]{baruch2021arkitscenes}
Gilad Baruch, Zhuoyuan Chen, Afshin Dehghan, Tal Dimry, Yuri Feigin, Peter Fu,
  Thomas Gebauer, Brandon Joffe, Daniel Kurz, Arik Schwartz, and Elad Shulman.
\newblock {ARKitScenes}: A diverse real-world dataset for {3D} indoor scene
  understanding using mobile {RGB-D} data.
\newblock In \emph{Proc. {NeurIPS} Datasets and Benchmarks}, 2021.

\bibitem[Bruce et~al.(2024)Bruce, Dennis, Edwards, Parker-Holder, Shi, Hughes,
  Lai, Mavalankar, Steigerwald, Apps, Aytar, Bechtle, Behbahani, Chan, Heess,
  Gonzalez, Osindero, Ozair, Reed, Zhang, Zolna, Clune, de~Freitas, Singh, and
  Rockt{\"a}schel]{bruce2024genie}
Jake Bruce, Michael~D. Dennis, Ashley Edwards, Jack Parker-Holder, Yuge Shi,
  Edward Hughes, Matthew Lai, Aditi Mavalankar, Richie Steigerwald, Chris Apps,
  Yusuf Aytar, Sarah Bechtle, Feryal M.~P. Behbahani, Stephanie C.~Y. Chan,
  Nicolas Heess, Lucy Gonzalez, Simon Osindero, Sherjil Ozair, Scott~E. Reed,
  Jingwei Zhang, Konrad Zolna, Jeff Clune, Nando de Freitas, Satinder Singh,
  and Tim Rockt{\"a}schel.
\newblock Genie: Generative interactive environments.
\newblock In \emph{Proc. {ICML}}, 2024.

\bibitem[Burgert et~al.(2025)Burgert, Xu, Xian, Pilarski, Clausen, He, Ma,
  Deng, Li, Mousavi, Ryoo, Debevec, and Yu]{burgert2025gowiththeflow}
Ryan Burgert, Yuancheng Xu, Wenqi Xian, Oliver Pilarski, Pascal Clausen,
  Mingming He, Li Ma, Yitong Deng, Lingxiao Li, Mohsen Mousavi, Michael Ryoo,
  Paul Debevec, and Ning Yu.
\newblock {Go-with-the-Flow}: Motion-controllable video diffusion models using
  real-time warped noise.
\newblock In \emph{Proc. {CVPR}}, 2025.

\bibitem[Carion et~al.(2026)Carion, Gustafson, Hu, Debnath, Hu, Suris, Ryali,
  Alwala, Khedr, Huang, Lei, Ma, Guo, Kalla, Marks, Greer, Wang, Sun,
  R{\"a}dle, Afouras, Mavroudi, Xu, Wu, Zhou, Momeni, Hazra, Ding, Vaze,
  Porcher, Li, Li, Kamath, Cheng, Doll{\'a}r, Ravi, Saenko, Zhang, and
  Feichtenhofer]{carion2026sam3}
Nicolas Carion, Laura Gustafson, Yuan-Ting Hu, Shoubhik Debnath, Ronghang Hu,
  Didac Suris, Chaitanya Ryali, Kalyan~Vasudev Alwala, Haitham Khedr, Andrew
  Huang, Jie Lei, Tengyu Ma, Baishan Guo, Arpit Kalla, Markus Marks, Joseph
  Greer, Meng Wang, Peize Sun, Roman R{\"a}dle, Triantafyllos Afouras,
  Effrosyni Mavroudi, Katherine Xu, Tsung-Han Wu, Yu Zhou, Liliane Momeni,
  Rishi Hazra, Shuangrui Ding, Sagar Vaze, Francois Porcher, Feng Li, Siyuan
  Li, Aishwarya Kamath, Ho~Kei Cheng, Piotr Doll{\'a}r, Nikhila Ravi, Kate
  Saenko, Pengchuan Zhang, and Christoph Feichtenhofer.
\newblock {SAM} 3: {Segment Anything} with concepts.
\newblock In \emph{Proc. {ICLR}}, 2026.

\bibitem[Chang et~al.(2019)Chang, Lambert, Sangkloy, Singh, Bak, Hartnett,
  Wang, Carr, Lucey, Ramanan, and Hays]{chang2019argoverse}
Ming-Fang Chang, John Lambert, Patsorn Sangkloy, Jagjeet Singh, Slawomir Bak,
  Andrew Hartnett, De Wang, Peter Carr, Simon Lucey, Deva Ramanan, and James
  Hays.
\newblock {Argoverse}: {3D} tracking and forecasting with rich maps.
\newblock In \emph{Proc. {CVPR}}, 2019.

\bibitem[Chao et~al.(2021)Chao, Yang, Xiang, Molchanov, Handa, Tremblay,
  Narang, Wyk, Iqbal, Birchfield, Kautz, and Fox]{chao2021dexycb}
Yu-Wei Chao, Wei Yang, Yu Xiang, Pavlo Molchanov, Ankur Handa, Jonathan
  Tremblay, Yashraj~S. Narang, Karl~Van Wyk, Umar Iqbal, Stan Birchfield, Jan
  Kautz, and Dieter Fox.
\newblock {DexYCB}: A benchmark for capturing hand grasping of objects.
\newblock In \emph{Proc. {CVPR}}, 2021.

\bibitem[Charatan et~al.(2024)Charatan, Li, Tagliasacchi, and
  Sitzmann]{charatan23pixelsplat}
David Charatan, Sizhe~Lester Li, Andrea Tagliasacchi, and Vincent Sitzmann.
\newblock {pixelSplat}: {3D} {Gaussian} splats from image pairs for scalable
  generalizable {3D} reconstruction.
\newblock In \emph{Proc. {CVPR}}, 2024.

\bibitem[Chen et~al.(2024{\natexlab{a}})Chen, Mons{\'o}, Du, Simchowitz,
  Tedrake, and Sitzmann]{chen24diffusion}
Boyuan Chen, Diego~Mart{\'i} Mons{\'o}, Yilun Du, Max Simchowitz, Russ Tedrake,
  and Vincent Sitzmann.
\newblock {Diffusion Forcing}: Next-token prediction meets full-sequence
  diffusion.
\newblock In \emph{Proc. {NeurIPS}}, 2024{\natexlab{a}}.

\bibitem[Chen et~al.(2025)Chen, Jiang, Liu, Gupta, Li, Zhao, and
  Wang]{chen2025physgen3d}
Boyuan Chen, Hanxiao Jiang, Shaowei Liu, Saurabh Gupta, Yunzhu Li, Hao Zhao,
  and Shenlong Wang.
\newblock {PhysGen3D}: Crafting a miniature interactive world from a single
  image.
\newblock In \emph{Proc. {CVPR}}, 2025.

\bibitem[Chen et~al.(2024{\natexlab{b}})Chen, Walsman, Memmel, Mo, Fang, Fox,
  and Gupta]{chen2024urdformer}
Qiuyu Chen, Aaron Walsman, Marius Memmel, Kaichun Mo, Alex Fang, Dieter Fox,
  and Abhishek Gupta.
\newblock {URDFormer}: A pipeline for constructing articulated simulation
  environments from real-world images.
\newblock In \emph{Proc. {RSS}}, 2024{\natexlab{b}}.

\bibitem[Chen et~al.(2026{\natexlab{a}})Chen, Chu, Gleize, Liang, Sax, Tang,
  Wang, Guo, Hardin, Li, Lin, Liu, Ma, Sagar, Song, Wang, Yang, Zhang,
  Doll{\'a}r, Gkioxari, Feiszli, and Malik]{chen2026sam3d}
Xingyu Chen, Fu-Jen Chu, Pierre Gleize, Kevin~J. Liang, Alexander Sax, Hao
  Tang, Weiyao Wang, Michelle Guo, Thibaut Hardin, Xiang Li, Aohan Lin, Jiawei
  Liu, Ziqi Ma, Anushka Sagar, Bowen Song, Xiaodong Wang, Jianing Yang, Bowen
  Zhang, Piotr Doll{\'a}r, Georgia Gkioxari, Matt Feiszli, and Jitendra Malik.
\newblock {SAM} {3D}: {3Dfy} anything in images.
\newblock In \emph{Proc. {CVPR}}, 2026{\natexlab{a}}.

\bibitem[Chen et~al.(2024{\natexlab{c}})Chen, Xu, Zheng, Zhuang, Pollefeys,
  Geiger, Cham, and Cai]{chen2024mvsplat}
Yuedong Chen, Haofei Xu, Chuanxia Zheng, Bohan Zhuang, Marc Pollefeys, Andreas
  Geiger, Tat-Jen Cham, and Jianfei Cai.
\newblock {MVSplat}: Efficient {3D} {Gaussian} splatting from sparse multi-view
  images.
\newblock In \emph{Proc. {ECCV}}, 2024{\natexlab{c}}.

\bibitem[Chen et~al.(2026{\natexlab{b}})Chen, Lin, and Zhang]{chen2026cwm}
Yiwen Chen, Guosheng Lin, and Chi Zhang.
\newblock {Code World Model}: Coding agent as world brain.
\newblock \emph{arXiv}, 2608.25927, 2026{\natexlab{b}}.

\bibitem[Chi et~al.(2026)Chi, Miao, Shi, Wu, Wang, Chen, Dai, Li, Ren, Gao,
  Long, Duan, Lyu, Wu, and Liu]{mirros2026agentgarten}
Jiawei Chi, Shangchen Miao, Zhiyuan Shi, Kailu Wu, Hanyang Wang, Weiliang Chen,
  Qiyu Dai, Chengyang Li, Jinshan Ren, Jun Gao, Mingsheng Long, Yueqi Duan,
  Jiangran Lyu, Jialong Wu, and Fangfu Liu.
\newblock {AgentGarten}: Code worlds for evolving agents.
\newblock \emph{arXiv}, 2610.12374, 2026.

\bibitem[Deitke et~al.(2022)Deitke, VanderBilt, Herrasti, Weihs, Salvador,
  Ehsani, Han, Kolve, Farhadi, Kembhavi, and Mottaghi]{deitke2022procthor}
Matt Deitke, Eli VanderBilt, Alvaro Herrasti, Luca Weihs, Jordi Salvador, Kiana
  Ehsani, Winson Han, Eric Kolve, Ali Farhadi, Aniruddha Kembhavi, and Roozbeh
  Mottaghi.
\newblock {ProcTHOR}: Large-scale embodied {AI} using procedural generation.
\newblock In \emph{Proc. {NeurIPS}}, 2022.

\bibitem[DeTone et~al.(2026)DeTone, Shen, Zhang, Ma, Straub, Newcombe, and
  Engel]{detone2026boxer}
Daniel DeTone, Tianwei Shen, Fan Zhang, Lingni Ma, Julian Straub, Richard
  Newcombe, and Jakob Engel.
\newblock {Boxer}: Robust lifting of open-world {2D} bounding boxes to {3D}.
\newblock In \emph{Proc. {ECCV}}, 2026.

\bibitem[Fang et~al.(2026)Fang, Shao, Zhu, Xiao, Chen, Liu, Fu, Cao, Zhai, and
  Zha]{fang2026coder}
Zixun Fang, Yawen Shao, Kai Zhu, Jie Xiao, Shihan Chen, Yu Liu, Xueyang Fu,
  Yang Cao, Wei Zhai, and Zheng-Jun Zha.
\newblock Code plans, diffusion renders: Open-ended generative world modeling.
\newblock \emph{arXiv}, 2609.26458, 2026.

\bibitem[Feng et~al.(2025)Feng, Zhang, Wang, Ye, Yu, Black, Darrell, and
  Kanazawa]{feng2025st4rtrack}
Haiwen Feng, Junyi Zhang, Qianqian Wang, Yufei Ye, Pengcheng Yu, Michael~J.
  Black, Trevor Darrell, and Angjoo Kanazawa.
\newblock {St4RTrack}: Simultaneous {4D} reconstruction and tracking in the
  world.
\newblock In \emph{Proc. {ICCV}}, 2025.

\bibitem[Foo et~al.(2026)Foo, Huang, Lattas, Moschoglou, Beeler, and
  Theobalt]{foo2026psivg}
Lin~Geng Foo, Mark~He Huang, Alexandros Lattas, Stylianos Moschoglou, Thabo
  Beeler, and Christian Theobalt.
\newblock Physical simulator in-the-loop video generation.
\newblock In \emph{Proc. {CVPR}}, 2026.

\bibitem[Gao et~al.(2024)Gao, Ho{\l}y{\'n}ski, Henzler, Brussee,
  Martin-Brualla, Srinivasan, Barron, and Poole]{gao2024cat3d}
Ruiqi Gao, Aleksander Ho{\l}y{\'n}ski, Philipp Henzler, Arthur Brussee, Ricardo
  Martin-Brualla, Pratul~P. Srinivasan, Jonathan~T. Barron, and Ben Poole.
\newblock {CAT3D}: Create anything in {3D} with multi-view diffusion models.
\newblock In \emph{Proc. {NeurIPS}}, 2024.

\bibitem[Ha and Schmidhuber(2018)]{ha2018worldmodels}
David Ha and J{\"u}rgen Schmidhuber.
\newblock Recurrent world models facilitate policy evolution.
\newblock In \emph{Proc. {NeurIPS}}, 2018.

\bibitem[Hafner et~al.(2025)Hafner, Pasukonis, Ba, and
  Lillicrap]{hafner2023dreamerv3}
Danijar Hafner, Jurgis Pasukonis, Jimmy Ba, and Timothy Lillicrap.
\newblock Mastering diverse control tasks through world models.
\newblock \emph{Nature}, 640\penalty0 (8059):\penalty0 647--653, 2025.

\bibitem[Heusel et~al.(2017)Heusel, Ramsauer, Unterthiner, Nessler, and
  Hochreiter]{heusel2017fid}
Martin Heusel, Hubert Ramsauer, Thomas Unterthiner, Bernhard Nessler, and Sepp
  Hochreiter.
\newblock {GANs} trained by a two time-scale update rule converge to a local
  {Nash} equilibrium.
\newblock In \emph{Proc. {NeurIPS}}, 2017.

\bibitem[Hu et~al.(2022)Hu, Shen, Wallis, Allen-Zhu, Li, Wang, Wang, and
  Chen]{hu2022lora}
Edward~J. Hu, Yelong Shen, Phillip Wallis, Zeyuan Allen-Zhu, Yuanzhi Li, Shean
  Wang, Lu Wang, and Weizhu Chen.
\newblock {LoRA}: Low-rank adaptation of large language models.
\newblock In \emph{Proc. {ICLR}}, 2022.

\bibitem[Huang et~al.(2025)Huang, Li, He, Zhou, and Shechtman]{huang25self}
Xun Huang, Zhengqi Li, Guande He, Mingyuan Zhou, and Eli Shechtman.
\newblock {Self Forcing}: Bridging the train-test gap in autoregressive video
  diffusion.
\newblock In \emph{Proc. {NeurIPS}}, 2025.

\bibitem[Huang et~al.(2026{\natexlab{a}})Huang, Li, Chiu, Lyu, Zhou, Yao,
  Lasenby, and Wu]{huang2026literealityagent}
Zhening Huang, Yueyan Li, Johnathan Chiu, Xiaoyang Lyu, Matt Zhou, Yuxin Yao,
  Joan Lasenby, and Shangzhe Wu.
\newblock {LiteReality-Agent}: An agentic system for interactable {3D} indoor
  scene reconstruction.
\newblock \emph{arXiv}, 2610.01863, 2026{\natexlab{a}}.

\bibitem[Huang et~al.(2026{\natexlab{b}})Huang, Lin, Lin, Huang, Yu, Niu, Yang,
  Liu, Chuang, Zhang, and Wang]{huang2026pwm}
Zheng-Hui Huang, Guixu Lin, Jiacheng Lin, Yi-Chuan Huang, Ruihan Yu, Muyao Niu,
  Siqi Yang, Yu-Lun Liu, Yung-Yu Chuang, Kaipeng Zhang, and Zhixiang Wang.
\newblock {Programmable World Model}.
\newblock \emph{arXiv}, 2609.10540, 2026{\natexlab{b}}.

\bibitem[Jiang et~al.(2025)Jiang, Mao, Xu, Lu, Ren, Jin, Xu, Yu, Pang, Zhao,
  Lin, and Dai]{jiang2025anysplat}
Lihan Jiang, Yucheng Mao, Linning Xu, Tao Lu, Kerui Ren, Yichen Jin, Xudong Xu,
  Mulin Yu, Jiangmiao Pang, Feng Zhao, Dahua Lin, and Bo Dai.
\newblock {AnySplat}: Feed-forward {3D} {Gaussian} splatting from unconstrained
  views.
\newblock \emph{ACM Trans. on Graphics (TOG)}, 44\penalty0 (6), 2025.

\bibitem[Karhade et~al.(2026)Karhade, Keetha, Zhang, Gupta, Sharma, Scherer,
  and Ramanan]{karhade2026any4d}
Jay Karhade, Nikhil Keetha, Yuchen Zhang, Tanisha Gupta, Akash Sharma,
  Sebastian Scherer, and Deva Ramanan.
\newblock {Any4D}: Unified feed-forward metric {4D} reconstruction.
\newblock In \emph{Proc. {CVPR}}, 2026.

\bibitem[Knapitsch et~al.(2017)Knapitsch, Park, Zhou, and
  Koltun]{knapitsch2017tanks}
Arno Knapitsch, Jaesik Park, Qian-Yi Zhou, and Vladlen Koltun.
\newblock {Tanks and Temples}: Benchmarking large-scale scene reconstruction.
\newblock \emph{ACM Trans. on Graphics (TOG)}, 36\penalty0 (4), 2017.

\bibitem[Lan et~al.(2026{\natexlab{a}})Lan, Jiang, Namekata, Pan, Zheng, and
  Vedaldi]{lan2026holo4d}
Yushi Lan, Zeren Jiang, Koichi Namekata, Xingang Pan, Chuanxia Zheng, and
  Andrea Vedaldi.
\newblock {Holo4D}: Holistic {4D} reconstruction as geometric control for video
  diffusion.
\newblock In \emph{Proc. {NeurIPS}}, 2026{\natexlab{a}}.

\bibitem[Lan et~al.(2026{\natexlab{b}})Lan, Luo, Hong, Zhou, Chen, Lyu, Dai,
  Yang, Loy, and Pan]{lan26stream3r}
Yushi Lan, Yihang Luo, Fangzhou Hong, Shangchen Zhou, Honghua Chen, Zhaoyang
  Lyu, Bo Dai, Shuai Yang, Chen~Change Loy, and Xingang Pan.
\newblock {STream3R}: Scalable sequential {3D} reconstruction with causal
  transformer.
\newblock In \emph{Proc. {ICLR}}, 2026{\natexlab{b}}.

\bibitem[Lazarow et~al.(2025)Lazarow, Griffiths, Kohavi, Crespo, and
  Dehghan]{lazarow2025cubify}
Justin Lazarow, David Griffiths, Gefen Kohavi, Francisco Crespo, and Afshin
  Dehghan.
\newblock {Cubify Anything}: Scaling indoor {3D} object detection.
\newblock In \emph{Proc. {CVPR}}, 2025.

\bibitem[LeCun(2022)]{lecun22a-path}
Yann LeCun.
\newblock A path towards autonomous machine intelligence.
\newblock OpenReview, \url{https://openreview.net/forum?id=BZ5a1r-kVsf}, 2022.

\bibitem[Li et~al.(2025{\natexlab{a}})Li, Michel, Pan, Liu, Roberts, and
  Xie]{li2025pisa}
Chenyu Li, Oscar Michel, Xichen Pan, Sainan Liu, Mike Roberts, and Saining Xie.
\newblock {PISA} experiments: Exploring physics post-training for video
  diffusion models by watching stuff drop.
\newblock In \emph{Proc. {ICML}}, 2025{\natexlab{a}}.

\bibitem[Li et~al.(2025{\natexlab{b}})Li, Torr, Vedaldi, and Jakab]{li25vmem}
Runjia Li, Philip Torr, Andrea Vedaldi, and Tomas Jakab.
\newblock {VMem}: Consistent interactive video scene generation with
  surfel-indexed view memory.
\newblock In \emph{Proc. {ICCV}}, 2025{\natexlab{b}}.

\bibitem[Li et~al.(2025{\natexlab{c}})Li, Yu, Liu, Yang, Herrmann, Wetzstein,
  and Wu]{li2025wonderplay}
Zizhang Li, Hong-Xing Yu, Wei Liu, Yin Yang, Charles Herrmann, Gordon
  Wetzstein, and Jiajun Wu.
\newblock {WonderPlay}: Dynamic {3D} scene generation from a single image and
  actions.
\newblock In \emph{Proc. {ICCV}}, 2025{\natexlab{c}}.

\bibitem[Liang et~al.(2026)Liang, Zhang, Qian, Chen, and
  Wan]{liang2026codevideomodel}
Zhiyang Liang, Jiawei Zhang, Guocheng Qian, Dongdong Chen, and Ziyu Wan.
\newblock {Code Video Model}.
\newblock Tech report,
  \url{https://code-video-model.github.io/tech_report.pdf}, 2026.

\bibitem[Liao et~al.(2026)Liao, Luo, Wu, Jin, Wu, Lin, Zhao, Wang, Li, and
  Loy]{liao2026puffin}
Kang Liao, Yihang Luo, Xiao-Ming Wu, Linyi Jin, Size Wu, Chunyu Lin, Yao Zhao,
  Fei Wang, Wei Li, and Chen~Change Loy.
\newblock {Puffin-World}: Scaling a unified multimodal model with native {3D}
  world states.
\newblock \emph{arXiv}, 2609.04196, 2026.

\bibitem[Lin et~al.(2026{\natexlab{a}})Lin, Chen, Liew, Chen, Li, Zhao, Peng,
  Guo, Zhou, Shi, Feng, and Kang]{lin2026da3}
Haotong Lin, Sili Chen, Jun~Hao Liew, Donny~Y. Chen, Zhenyu Li, Yang Zhao, Sida
  Peng, Hengkai Guo, Xiaowei Zhou, Guang Shi, Jiashi Feng, and Bingyi Kang.
\newblock {Depth Anything 3}: Recovering the visual space from any views.
\newblock In \emph{Proc. {ICLR}}, 2026{\natexlab{a}}.

\bibitem[Lin et~al.(2026{\natexlab{b}})Lin, Liu, Salamanca, Kant, Burgert, Xu,
  Namekata, Zhao, Zhou, Goldblum, Debevec, and Yu]{lin2026vista4d}
Kuan~Heng Lin, Zhizheng Liu, Pablo Salamanca, Yash Kant, Ryan Burgert,
  Yuancheng Xu, Koichi Namekata, Yiwei Zhao, Bolei Zhou, Micah Goldblum, Paul
  Debevec, and Ning Yu.
\newblock {Vista4D}: Video reshooting with {4D} point clouds.
\newblock In \emph{Proc. {CVPR}}, 2026{\natexlab{b}}.

\bibitem[Lipman et~al.(2023)Lipman, Chen, Ben-Hamu, Nickel, and
  Le]{lipman2023flow}
Yaron Lipman, Ricky T.~Q. Chen, Heli Ben-Hamu, Maximilian Nickel, and Matthew
  Le.
\newblock Flow matching for generative modeling.
\newblock In \emph{Proc. {ICLR}}, 2023.

\bibitem[Liu et~al.(2024)Liu, Ren, Gupta, and Wang]{liu2024physgen}
Shaowei Liu, Zhongzheng Ren, Saurabh Gupta, and Shenlong Wang.
\newblock {PhysGen}: Rigid-body physics-grounded image-to-video generation.
\newblock In \emph{Proc. {ECCV}}, 2024.

\bibitem[Liu et~al.(2026{\natexlab{a}})Liu, Chen, Li, Wang, Yu, and
  Wu]{liu2026realwonder}
Wei Liu, Ziyu Chen, Zizhang Li, Yue Wang, Hong-Xing Yu, and Jiajun Wu.
\newblock {RealWonder}: Real-time physical action-conditioned video generation.
\newblock \emph{arXiv}, 2603.05449, 2026{\natexlab{a}}.

\bibitem[Liu et~al.(2026{\natexlab{b}})Liu, Xiao, Chen, Feng, Tai, Tang, and
  Kang]{liu2026traceanything}
Xinhang Liu, Yuxi Xiao, Donny~Y. Chen, Jiashi Feng, Yu-Wing Tai, Chi-Keung
  Tang, and Bingyi Kang.
\newblock {Trace Anything}: Representing any video in {4D} via trajectory
  fields.
\newblock In \emph{Proc. {ICLR}}, 2026{\natexlab{b}}.

\bibitem[Loshchilov and Hutter(2019)]{loshchilov19decoupled}
Ilya Loshchilov and Frank Hutter.
\newblock Decoupled weight decay regularization.
\newblock In \emph{Proc. {ICLR}}, 2019.

\bibitem[Lu et~al.(2026)Lu, Ye, Li, Xue, Liu, Han, and
  Xie]{lu2026streamrenderh3}
Yanzuo Lu, Tian Ye, Yitong Li, Shuchen Xue, Haozhe Liu, Song Han, and Enze Xie.
\newblock {StreamRender-H3}: Interactive coding worlds with real-time streaming
  diffusion render.
\newblock \url{https://nvlabs.github.io/Sana/Sol-Engine/StreamRender-H3/},
  2026.

\bibitem[Luo et~al.(2026)Luo, Zhou, Lan, Pan, and Loy]{luo20264rc}
Yihang Luo, Shangchen Zhou, Yushi Lan, Xingang Pan, and Chen~Change Loy.
\newblock {4RC}: {4D} reconstruction via conditional querying anytime and
  anywhere.
\newblock In \emph{Proc. {ICML}}, 2026.

\bibitem[Mandi et~al.(2025)Mandi, Weng, Bauer, and Song]{mandi2024real2code}
Zhao Mandi, Yijia Weng, Dominik Bauer, and Shuran Song.
\newblock {Real2Code}: Reconstruct articulated objects via code generation.
\newblock In \emph{Proc. {ICLR}}, 2025.

\bibitem[Mao et~al.(2025)Mao, Zhong, Fang, Zheng, Tang, Zhu, Tan, and
  Zhou]{mao2025spatiallm}
Yongsen Mao, Junhao Zhong, Chuan Fang, Jia Zheng, Rui Tang, Hao Zhu, Ping Tan,
  and Zihan Zhou.
\newblock {SpatialLM}: Training large language models for structured indoor
  modeling.
\newblock In \emph{Proc. {NeurIPS}}, 2025.

\bibitem[Mazur et~al.(2026)Mazur, Karaev, Chang, Malik, and
  Shafiullah]{mazur2026agentstar}
Kirill Mazur, Nikita Karaev, Matthew Chang, Jitendra Malik, and Nur
  Muhammad~Mahi Shafiullah.
\newblock {AgentSTAR}: Agentic shape tracking and reconstruction from monocular
  videos.
\newblock \emph{arXiv}, 2609.24487, 2026.

\bibitem[{MiniMax}(2026)]{minimax2026h3}
{MiniMax}.
\newblock {MiniMax H3}: An open model breaking the boundaries between tasks and
  modalities.
\newblock \url{https://minimax.io/blog/minimax-h3}, 2026.

\bibitem[{NVIDIA}(2025)]{nvidia2025cosmos}
{NVIDIA}.
\newblock {Cosmos} world foundation model platform for physical {AI}.
\newblock \emph{arXiv}, 2501.03575, 2025.

\bibitem[{OpenAI}(2026{\natexlab{a}})]{openai2026gpt56}
{OpenAI}.
\newblock {GPT-5.6} system card.
\newblock \url{https://deploymentsafety.openai.com/gpt-5-6},
  2026{\natexlab{a}}.

\bibitem[{OpenAI}(2026{\natexlab{b}})]{openai2026gpt6astra}
{OpenAI}.
\newblock {GPT-6 Astra} system card.
\newblock \url{https://deploymentsafety.openai.com/gpt-6-astra},
  2026{\natexlab{b}}.

\bibitem[Pan et~al.(2023)Pan, Charron, Yang, Peters, Whelan, Kong, Parkhi,
  Newcombe, and Ren]{pan2023adt}
Xiaqing Pan, Nicholas Charron, Yongqian Yang, Scott Peters, Thomas Whelan, Chen
  Kong, Omkar Parkhi, Richard Newcombe, and Carl~Yuheng Ren.
\newblock {Aria Digital Twin}: A new benchmark dataset for egocentric {3D}
  machine perception.
\newblock In \emph{Proc. {ICCV}}, 2023.

\bibitem[Ravi et~al.(2025)Ravi, Gabeur, Hu, Hu, Ryali, Ma, Khedr, R{\"a}dle,
  Rolland, Gustafson, Mintun, Pan, Alwala, Carion, Wu, Girshick, Doll{\'a}r,
  and Feichtenhofer]{ravi2025sam2}
Nikhila Ravi, Valentin Gabeur, Yuan-Ting Hu, Ronghang Hu, Chaitanya Ryali,
  Tengyu Ma, Haitham Khedr, Roman R{\"a}dle, Chlo{\'e} Rolland, Laura
  Gustafson, Eric Mintun, Junting Pan, Kalyan~Vasudev Alwala, Nicolas Carion,
  Chao-Yuan Wu, Ross~B. Girshick, Piotr Doll{\'a}r, and Christoph
  Feichtenhofer.
\newblock {SAM} 2: {Segment Anything} in images and videos.
\newblock In \emph{Proc. {ICLR}}, 2025.

\bibitem[Ren et~al.(2025)Ren, Shen, Huang, Ling, Lu, Nimier-David, M{\"u}ller,
  Keller, Fidler, and Gao]{ren2025gen3c}
Xuanchi Ren, Tianchang Shen, Jiahui Huang, Huan Ling, Yifan Lu, Merlin
  Nimier-David, Thomas M{\"u}ller, Alexander Keller, Sanja Fidler, and Jun Gao.
\newblock {GEN3C}: {3D}-informed world-consistent video generation with precise
  camera control.
\newblock In \emph{Proc. {CVPR}}, 2025.

\bibitem[{Robbyant Team}(2026)]{robbyant2026lingbotworld}
{Robbyant Team}.
\newblock Advancing open-source world models.
\newblock \emph{arXiv}, 2601.20540, 2026.

\bibitem[Song et~al.(2025)Song, Chen, Simchowitz, Du, Tedrake, and
  Sitzmann]{song25history-guided}
Kiwhan Song, Boyuan Chen, Max Simchowitz, Yilun Du, Russ Tedrake, and Vincent
  Sitzmann.
\newblock History-guided video diffusion.
\newblock In \emph{Proc. {ICML}}, 2025.

\bibitem[Sucar et~al.(2026)Sucar, Insafutdinov, Lai, and
  Vedaldi]{sucar2026vdpm}
Edgar Sucar, Eldar Insafutdinov, Zihang Lai, and Andrea Vedaldi.
\newblock {V-DPM}: {4D} video reconstruction with dynamic point maps.
\newblock In \emph{Proc. {CVPR}}, 2026.

\bibitem[Todorov et~al.(2012)Todorov, Erez, and Tassa]{todorov2012mujoco}
Emanuel Todorov, Tom Erez, and Yuval Tassa.
\newblock {MuJoCo}: A physics engine for model-based control.
\newblock In \emph{Proc. {IROS}}, 2012.

\bibitem[{Unitree Robotics}(2024)]{unitree2024rlgym}
{Unitree Robotics}.
\newblock {Unitree RL GYM}.
\newblock \url{https://github.com/unitreerobotics/unitree_rl_gym}, 2024.

\bibitem[Valevski et~al.(2025)Valevski, Leviathan, Arar, and
  Fruchter]{valevski2024gamengen}
Dani Valevski, Yaniv Leviathan, Moab Arar, and Shlomi Fruchter.
\newblock Diffusion models are real-time game engines.
\newblock In \emph{Proc. {ICLR}}, 2025.

\bibitem[Villasevil et~al.(2024)Villasevil, Simeonov, Li, Chan, Chen, Gupta,
  and Agrawal]{torne2024rialto}
Marcel~Torne Villasevil, Anthony Simeonov, Zechu Li, April Chan, Tao Chen,
  Abhishek Gupta, and Pulkit Agrawal.
\newblock Reconciling reality through simulation: A real-to-sim-to-real
  approach for robust manipulation.
\newblock In \emph{Proc. {RSS}}, 2024.

\bibitem[Wan et~al.(2025)Wan, Wang, Ai, Wen, Mao, Xie, Chen, Yu, Zhao, Yang,
  Zeng, Wang, Zhang, Zhou, Wang, Chen, Zhu, Zhao, Yan, Huang, Feng, Zhang, Li,
  Wu, Chu, Feng, Zhang, Sun, Fang, Wang, Gui, Weng, Shen, Lin, Wang, Wang,
  Zhou, Wang, Shen, Yu, Shi, Huang, Xu, Kou, Lv, Li, Liu, Wang, Zhang, Huang,
  Li, Wu, Liu, Pan, Zheng, Hong, Shi, Feng, Jiang, Han, Wu, and
  Liu]{wan2025wan}
Team Wan, Ang Wang, Baole Ai, Bin Wen, Chaojie Mao, Chen-Wei Xie, Di Chen,
  Feiwu Yu, Haiming Zhao, Jianxiao Yang, Jianyuan Zeng, Jiayu Wang, Jingfeng
  Zhang, Jingren Zhou, Jinkai Wang, Jixuan Chen, Kai Zhu, Kang Zhao, Keyu Yan,
  Lianghua Huang, Mengyang Feng, Ningyi Zhang, Pandeng Li, Pingyu Wu, Ruihang
  Chu, Ruili Feng, Shiwei Zhang, Siyang Sun, Tao Fang, Tianxing Wang, Tianyi
  Gui, Tingyu Weng, Tong Shen, Wei Lin, Wei Wang, Wei Wang, Wenmeng Zhou, Wente
  Wang, Wenting Shen, Wenyuan Yu, Xianzhong Shi, Xiaoming Huang, Xin Xu, Yan
  Kou, Yangyu Lv, Yifei Li, Yijing Liu, Yiming Wang, Yingya Zhang, Yitong
  Huang, Yong Li, You Wu, Yu Liu, Yulin Pan, Yun Zheng, Yuntao Hong, Yupeng
  Shi, Yutong Feng, Zeyinzi Jiang, Zhen Han, Zhi-Fan Wu, and Ziyu Liu.
\newblock Wan: Open and advanced large-scale video generative models.
\newblock \emph{arXiv}, 2503.20314, 2025.

\bibitem[Wang et~al.(2025{\natexlab{a}})Wang, Chen, Huang, Dou, Liu, Gu, and
  Liu]{wang2025physctrl}
Chen Wang, Chuhao Chen, Yiming Huang, Zhiyang Dou, Yuan Liu, Jiatao Gu, and
  Lingjie Liu.
\newblock {PhysCtrl}: Generative physics for controllable and physics-grounded
  video generation.
\newblock In \emph{Proc. {NeurIPS}}, 2025{\natexlab{a}}.

\bibitem[Wang et~al.(2026{\natexlab{a}})Wang, Xie, Feng, Lv, Wang, and
  Jiang]{wang26streamforce}
Hanhui Wang, Yiming Xie, Haiwen Feng, Zhaoyang Lv, Shenlong Wang, and Huaizu
  Jiang.
\newblock Streaming video generation with streaming force control.
\newblock \emph{arXiv}, 2606.07508, 2026{\natexlab{a}}.

\bibitem[Wang et~al.(2025{\natexlab{b}})Wang, Chen, Karaev, Vedaldi, Rupprecht,
  and Novotn{\'y}]{wang2025vggt}
Jianyuan Wang, Minghao Chen, Nikita Karaev, Andrea Vedaldi, Christian
  Rupprecht, and David Novotn{\'y}.
\newblock {VGGT}: Visual geometry grounded transformer.
\newblock In \emph{Proc. {CVPR}}, 2025{\natexlab{b}}.

\bibitem[Wang et~al.(2026{\natexlab{b}})Wang, Zhou, Zhu, Chang, Zhou, Li, Chen,
  Pang, Shen, and He]{wang2026pi3}
Yifan Wang, Jianjun Zhou, Haoyi Zhu, Wenzheng Chang, Yang Zhou, Zizun Li, Junyi
  Chen, Jiangmiao Pang, Chunhua Shen, and Tong He.
\newblock $\pi^3$: Permutation-equivariant visual geometry learning.
\newblock In \emph{Proc. {ICLR}}, 2026{\natexlab{b}}.

\bibitem[Wiedemer et~al.(2025)Wiedemer, Li, Vicol, Gu, Matarese, Swersky, Kim,
  Jaini, and Geirhos]{wiedemer25video}
Thadd{\"a}us Wiedemer, Yuxuan Li, Paul Vicol, Shixiang~Shane Gu, Nick Matarese,
  Kevin Swersky, Been Kim, Priyank Jaini, and Robert Geirhos.
\newblock Video models are zero-shot learners and reasoners.
\newblock \emph{arXiv}, 2509.20328, 2025.

\bibitem[Wu et~al.(2026)Wu, Vuong, Jeon, Narasimhan, and
  Ramanan]{wu2026framecrafter}
Qi Wu, Khiem Vuong, Minsik Jeon, Srinivasa Narasimhan, and Deva Ramanan.
\newblock {FrameCrafter}: Novel view synthesis as video completion.
\newblock In \emph{Proc. {ECCV}}, 2026.

\bibitem[Wu et~al.(2024)Wu, Mildenhall, Henzler, Park, Gao, Watson, Srinivasan,
  Verbin, Barron, Poole, and Ho{\l}y{\'n}ski]{wu2023reconfusion}
Rundi Wu, Ben Mildenhall, Philipp Henzler, Keunhong Park, Ruiqi Gao, Daniel
  Watson, Pratul~P. Srinivasan, Dor Verbin, Jonathan~T. Barron, Ben Poole, and
  Aleksander Ho{\l}y{\'n}ski.
\newblock {ReconFusion}: {3D} reconstruction with diffusion priors.
\newblock In \emph{Proc. {CVPR}}, 2024.

\bibitem[Wu et~al.(2025)Wu, Yang, Po, Xu, Liu, Lin, and Wetzstein]{wu25video}
Tong Wu, Shuai Yang, Ryan Po, Yinghao Xu, Ziwei Liu, Dahua Lin, and Gordon
  Wetzstein.
\newblock Video world models with long-term spatial memory.
\newblock In \emph{Proc. {NeurIPS}}, 2025.

\bibitem[Xiang et~al.(2025)Xiang, Lv, Xu, Deng, Wang, Zhang, Chen, Tong, and
  Yang]{xiang2025trellis}
Jianfeng Xiang, Zelong Lv, Sicheng Xu, Yu Deng, Ruicheng Wang, Bowen Zhang,
  Dong Chen, Xin Tong, and Jiaolong Yang.
\newblock Structured {3D} latents for scalable and versatile {3D} generation.
\newblock In \emph{Proc. {CVPR}}, 2025.

\bibitem[Xiao et~al.(2025{\natexlab{a}})Xiao, Wang, Xue, Karaev, Makarov, Kang,
  Zhu, Bao, Shen, and Zhou]{xiao2025spatialtrackerv2}
Yuxi Xiao, Jianyuan Wang, Nan Xue, Nikita Karaev, Yuri Makarov, Bingyi Kang,
  Xing Zhu, Hujun Bao, Yujun Shen, and Xiaowei Zhou.
\newblock {SpatialTrackerV2}: Advancing {3D} point tracking with explicit
  camera motion.
\newblock In \emph{Proc. {ICCV}}, 2025{\natexlab{a}}.

\bibitem[Xiao et~al.(2025{\natexlab{b}})Xiao, Lan, Zhou, Ouyang, Yang, Zeng,
  and Pan]{xiao2026worldmem}
Zeqi Xiao, Yushi Lan, Yifan Zhou, Wenqi Ouyang, Shuai Yang, Yanhong Zeng, and
  Xingang Pan.
\newblock {WorldMem}: Long-term consistent world simulation with memory.
\newblock In \emph{Proc. {NeurIPS}}, 2025{\natexlab{b}}.

\bibitem[Xie et~al.(2024)Xie, Zong, Qiu, Li, Feng, Yang, and
  Jiang]{xie2024physgaussian}
Tianyi Xie, Zeshun Zong, Yuxing Qiu, Xuan Li, Yutao Feng, Yin Yang, and
  Chenfanfu Jiang.
\newblock {PhysGaussian}: Physics-integrated {3D} {Gaussians} for generative
  dynamics.
\newblock In \emph{Proc. {CVPR}}, 2024.

\bibitem[Yeshwanth et~al.(2023)Yeshwanth, Liu, Nie{\ss}ner, and
  Dai]{yeshwanth2023scannetpp}
Chandan Yeshwanth, Yueh-Cheng Liu, Matthias Nie{\ss}ner, and Angela Dai.
\newblock {ScanNet}++: A high-fidelity dataset of {3D} indoor scenes.
\newblock In \emph{Proc. {ICCV}}, 2023.

\bibitem[Yu et~al.(2025{\natexlab{a}})Yu, Bai, Qin, Liu, Wang, Wan, Zhang, and
  Liu]{yu2025cam}
Jiwen Yu, Jianhong Bai, Yiran Qin, Quande Liu, Xintao Wang, Pengfei Wan, Di
  Zhang, and Xihui Liu.
\newblock {Context as Memory}: Scene-consistent interactive long video
  generation with memory retrieval.
\newblock In \emph{Proc. SIGGRAPH Asia}, 2025{\natexlab{a}}.

\bibitem[Yu et~al.(2025{\natexlab{b}})Yu, Qin, Wang, Wan, Zhang, and
  Liu]{yu25gamefactory}
Jiwen Yu, Yiran Qin, Xintao Wang, Pengfei Wan, Di Zhang, and Xihui Liu.
\newblock {GameFactory}: Creating new games with generative interactive videos.
\newblock In \emph{Proc. {ICCV}}, 2025{\natexlab{b}}.

\bibitem[Yu et~al.(2025{\natexlab{c}})Yu, Hu, Xing, and
  Shan]{yu2025trajectorycrafter}
Mark Yu, Wenbo Hu, Jinbo Xing, and Ying Shan.
\newblock {TrajectoryCrafter}: Redirecting camera trajectory for monocular
  videos via diffusion models.
\newblock In \emph{Proc. {ICCV}}, 2025{\natexlab{c}}.

\bibitem[Yu et~al.(2025{\natexlab{d}})Yu, Xing, Yuan, Hu, Li, Huang, Gao, Wong,
  Shan, and Tian]{yu2024viewcrafter}
Wangbo Yu, Jinbo Xing, Li Yuan, Wenbo Hu, Xiaoyu Li, Zhipeng Huang, Xiangjun
  Gao, Tien-Tsin Wong, Ying Shan, and Yonghong Tian.
\newblock {ViewCrafter}: Taming video diffusion models for high-fidelity novel
  view synthesis.
\newblock \emph{{IEEE} Trans. on Pattern Analysis and Machine Intelligence
  ({TPAMI})}, 2025{\natexlab{d}}.

\bibitem[Zhan et~al.(2026)Zhan, Li, Yu, and Wu]{zhan2026perpetualwonder}
Jiahao Zhan, Zizhang Li, Hong-Xing Yu, and Jiajun Wu.
\newblock {PerpetualWonder}: Long-horizon action-conditioned {4D} scene
  generation.
\newblock In \emph{Proc. {CVPR}}, 2026.

\bibitem[Zhang et~al.(2025)Zhang, Cai, Li, Wetzstein, and
  Agrawala]{zhang2025framepack}
Lvmin Zhang, Shengqu Cai, Muyang Li, Gordon Wetzstein, and Maneesh Agrawala.
\newblock Frame context packing and drift prevention in next-frame-prediction
  video diffusion models.
\newblock In \emph{Proc. {NeurIPS}}, 2025.

\bibitem[Zhang et~al.(2018)Zhang, Isola, Efros, Shechtman, and
  Wang]{zhang2018lpips}
Richard Zhang, Phillip Isola, Alexei~A. Efros, Eli Shechtman, and Oliver Wang.
\newblock The unreasonable effectiveness of deep features as a perceptual
  metric.
\newblock In \emph{Proc. {CVPR}}, 2018.

\bibitem[Zhang et~al.(2024)Zhang, Yu, Wu, Feng, Zheng, Snavely, Wu, and
  Freeman]{zhang2024physdreamer}
Tianyuan Zhang, Hong-Xing Yu, Rundi Wu, Brandon~Y. Feng, Changxi Zheng, Noah
  Snavely, Jiajun Wu, and William~T. Freeman.
\newblock {PhysDreamer}: Physics-based interaction with {3D} objects via video
  generation.
\newblock In \emph{Proc. {ECCV}}, 2024.

\bibitem[Zhou et~al.(2025)Zhou, Gao, Voleti, Vasishta, Yao, Boss, Torr,
  Rupprecht, and Jampani]{zhou2025seva}
Jensen Zhou, Hang Gao, Vikram Voleti, Aaryaman Vasishta, Chun-Han Yao, Mark
  Boss, Philip Torr, Christian Rupprecht, and Varun Jampani.
\newblock {Stable Virtual Camera}: Generative view synthesis with diffusion
  models.
\newblock In \emph{Proc. {ICCV}}, 2025.

\end{thebibliography}
}

\clearpage
\maketitlesupplementary
\appendix
\makeatletter\setlength{\@fptop}{0pt}\makeatother
\raggedbottom
\renewcommand{\dblfloatpagefraction}{0.85}

\noindent In this appendix, we first give more implementation details of \NICKNAME{} on agentic reconstruction, the scene program, and the renderer in \cref{sec:supp_impl}, and describe how the training data of the renderer are built in \cref{sec:supp_data}.
We then detail the evaluation in \cref{sec:supp_eval} and present additional results in \cref{sec:supp_comp}, including camera-controlled video generation with different numbers of context views, more qualitative comparisons with baselines, and control over physical properties.
In \cref{sec:supp_apps}, we show more applications enabled by the explicit proxy world, including spatial exploration, long takes, embodied simulation, bullet time, and dense 3D tracking.
We also provide video results on our project page, as introduced in \cref{sec:supp_video}.
Finally, we discuss limitations and future work in \cref{sec:supp_limit}.

\section{Implementation Details}
\label{sec:supp_impl}

\subsection{Agentic Reconstruction}
\label{sec:supp_impl_recon}

\paragraph{Agent.}
Unless otherwise specified, our experiments use GPT-6 Astra~\cite{openai2026gpt6astra} with the Codex harness as the agent.
It accesses the scene only through the tools described below.

\paragraph{Proposals.}
The agent first lists the object categories visible in the context views and marks the floor in them, which we use with the depth to orient the scene upright.
Boxer~\cite{detone2026boxer}, prompted with these categories, predicts the initial 3D box proposals.

\paragraph{Composition.}
For objects that the detector misses, SAM~3~\cite{carion2026sam3} is prompted with the same categories.
Its masks are lifted to 3D points with the depth, associated across views, and enclosed by upright boxes, and the boxes that no Boxer box covers are added as candidates.
In a single call without tools, the agent then composes the scene program from the context views and a bird's-eye view (BEV) of all candidates.
It keeps, merges, or omits every candidate, and adds a missing object only with a box and the views that show it as evidence.
It also writes the layout, \ie, the floors, walls, and ceilings of indoor spaces and the ground and open sky of outdoor ones, together with structures such as columns, stairs, and buildings.
The ground and the other surfaces of the layout are fitted to the depth, before and after each editing session.

\paragraph{Editing.}
The editing agent sees the proxy rendered from the context cameras and overlaid on the views, with one color per object, and a list of boxes that disagree with their SAM~3 masks.
Read-only tools render overlays at chosen views, enlarge a single object, and query the depth and 3D points at given pixels.
Editing tools modify the scene program, refit a box to the 3D points of its masks, fit a wall, floor, or ceiling to marked pixels, and solve \cref{eq:fit} of the main paper for one box.
Each edit returns the overlays before and after it, so the agent can verify its effect.
For \cref{eq:fit}, the agent marks two points on each chosen box edge in the views that show it, $d$ sums the distances of these points to the projected edge, and $\rho$ is the soft-$\ell_1$ loss.
The update is bounded to small changes of position, size, and yaw.
The fit is accepted only if its error in the remaining views that show the edge increases by at most half a pixel.

\paragraph{Review.}
The reviewer is a separate call with a fresh context.
It receives the overlays, a BEV, and the scene program, and can render further views with the read-only tools.
It either accepts the scene or returns issues, each tied to an object and the views that show it, for the next editing session.
By default in our experiments, each editing session is limited to 30 tool calls and 300\,s, and the loop ends when the reviewer accepts the scene, when a session leaves it unchanged, or after 64 tool calls in total.

\paragraph{Prompts.}
The prompts of the planning, composition, and review calls are abridged below.

\begin{promptbox}{Agent prompts (abridged)}
\textbf{Planning}
\begin{Verbatim}
Inspect all RGB input views and plan the vocabulary for open-vocabulary instance detection. [...]
- Return concise distinct physical object categories. Include visible main furniture, doors, windows, and user-requested objects. Avoid tiny clutter unless requested.
- Select at least six pixels on the real walking floor, distributed across views and across the floor area. Under the open sky, they lie on the open ground people walk or drive on. [...]
\end{Verbatim}
\smallskip
\textbf{Composition}
\begin{Verbatim}
Understand the whole scene and construct the initial complete program. [...]
- Proposals are boxes fitted to 3D points of detected instances. They are hypotheses: one physical object may arrive as several fragments (merge all their IDs into the single object), and a proposal can be spurious (omit it with a reason). Start every kept object from its proposal box and change only what the RGB contradicts.
- Things that are small and crowded together (books on a shelf, objects on a table, rows of seats) become one box for the group, or are left out when the group does not matter at the scale of the place. [...]
\end{Verbatim}
\smallskip
\textbf{Review}
\begin{Verbatim}
Review the current scene independently, without editing it. The attached images show the exported geometry as instance-colored RGB overlays. [...]
- Judge semantic completeness, object scope and placement, free space, glass, doors, walls, and RGB alignment. Numbers are diagnostics, not ground truth.
- Do not equate correct counts, separate instances, or approximate placement with alignment. Do not accept a scene just because compilation or numerical checks passed. [...]
\end{Verbatim}
\end{promptbox}

\subsection{Scene Program and Physics Engine}
\label{sec:supp_impl_physics}

\NICKNAME{} represents the proxy world as a scene program, which we execute in MuJoCo~\cite{todorov2012mujoco}.
It is written through Python calls as in \cref{fig:pipeline} of the main paper, each adding an entry to the stored JSON program excerpted below.
It stores the layout and every object with its category, the proposals it merges, and an oriented box in meters.
For an interaction, it further stores the mass, friction, and restitution of the moving objects and the timed actions, \eg, an initial velocity, an applied force, or a change of a global setting.
To execute the program, the agent writes scripts that compile it into a MuJoCo model, where the layout and static objects become fixed colliders and each moving object a rigid body with its stored properties.
Stepping this model yields the states $S_{1:T}$, which are rendered into the proxy video.
As described in \cref{sec:method_code} of the main paper, objects in contact are upgraded from boxes to finer geometry.
The agent composes them from primitive parts fitted to their silhouettes in the context views, \eg, a chair, or imports simulation-ready assets, \eg, the manipulators in \cref{fig:supp_robots}.

\begin{promptbox}{Scene program (excerpt)}
\begin{Verbatim}
{"layout": [{"id": "kitchen", "floor_height": 0,
   "ceiling_height": 2.37, "outer": [[-1.97, -0.17],
   [2.12, 0.93], [1.55, 3.09], [-2.55, 1.99]], ...}],
 "objects": [{"id": "cabinet_tall",
   "category": "kitchen cabinet", "source_ids":
   ["boxer_0046", "sam3_0078", "sam3_0079"],
   "box": {"translation": [-1.76, 0.25, 1.13],
     "rotation": [0.991, 0, 0, 0.131],
     "scale": [0.60, 0.64, 2.26]},
   "reason": "The apparent television proposal is
     glossy tall cabinetry."}, ...],
 "omissions": [{"source_ids":
   ["boxer_0016", "boxer_0103"], "reason":
   "Displaced duplicate hypotheses for plant_sink"}]}
\end{Verbatim}
\vspace{2pt}\hrule height 0.3pt\vspace{6.7pt}
\begin{Verbatim}
{"moving": [{"id": "chair_1", "mass": 5.0,
   "friction": 0.3, "restitution": 0.25,
   "v0": [0.18, -0.68, 0.0]}],
 "gravity_events": [{"t": 0.3, "g": [0, 0, 1.2]},
   {"t": 2.0, "g": [0, 0, -9.81]}]}
\end{Verbatim}
\end{promptbox}

\subsection{Renderer}
\label{sec:supp_impl_render}

\paragraph{Training.}
We freeze MiniMax H3 Ref2VA~\cite{minimax2026h3} and train LoRA~\cite{hu2022lora} of rank 128 on the attention and MLP layers of all its transformer blocks.
Since our training data contain no audio, we fine-tune it on video only with the flow-matching objective~\cite{lipman2023flow}, using 124-frame clips at $640\times480$.
We train for 3K iterations with AdamW~\cite{loshchilov19decoupled} and a learning rate of $5\times10^{-5}$.
Training runs on 56 NVIDIA A100 GPUs with 4-way tensor parallelism and 4-step gradient accumulation, which gives an effective batch size of 56.

\paragraph{Inference.}
In our experiments, we use 32 denoising steps without classifier-free guidance.
Qwen3-VL-8B~\cite{bai2025qwen3vl} writes the scene description of the text prompt from the context views and the objects visible in the proxy video.
With tensor parallelism on 8 NVIDIA A100 GPUs, generating a 124-frame video takes about two minutes.

\begin{figure}[t]
\centering
\begin{tikzpicture}[x=0.62pt, y=1pt, font=\scriptsize]
  \definecolor{chunkfill}{RGB}{214,222,235}
  \definecolor{fixedfill}{RGB}{120,140,175}
  \foreach \w/\lab in {0/1, 1/2, 2/3} {
    \pgfmathsetmacro{\s}{90*\w}
    \pgfmathsetmacro{\yy}{-12*\w}
    \fill[chunkfill] (\s, \yy) rectangle ++(124, 8);
    \ifnum\w>0
      \fill[fixedfill] (\s, \yy) rectangle ++(34, 8);
    \fi
    \draw[black!60, line width=0.3pt] (\s, \yy) rectangle ++(124, 8);
    \node[anchor=east] at (\s - 2, \yy + 4) {chunk \lab};
  }
  \draw[black!60, line width=0.3pt] (0, 11) -- (0, 15) -- (124, 15) -- (124, 11);
  \node[anchor=south] at (55, 15) {124 frames};
  \draw[black!60, line width=0.3pt] (90, 11) -- (90, 15);
  \node[anchor=south] at (107, 15) {34};
  \fill[fixedfill] (150, 1) rectangle ++(10, 6);
  \node[anchor=west] at (162, 4) {continuation context};
\end{tikzpicture}
\caption{\textbf{Long video inference with overlapping chunks.} \NICKNAME{} generates a long video in consecutive 124-frame chunks that overlap by 34 frames, so each new chunk advances the video by 90 frames. The last 34 generated frames of a chunk serve as the continuation context of the next.}
\label{fig:supp_chunks}
\vspace{-2mm}
\end{figure}
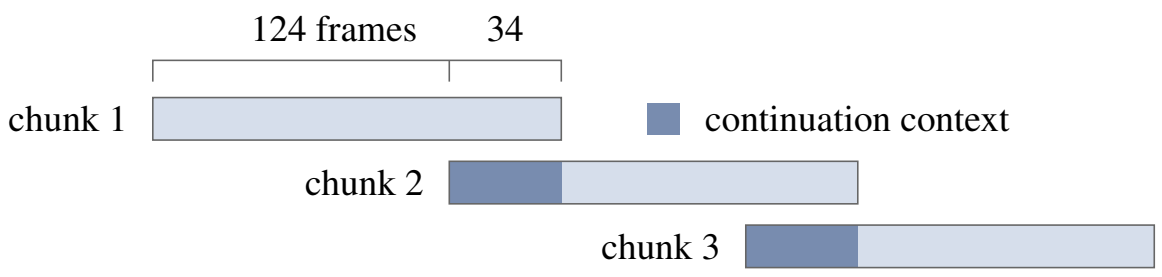

\paragraph{Long video inference.}
\label{sec:supp_impl_long}
\NICKNAME{} can also generate videos far longer than a single pass of the renderer.
Since the 3D proxy world persists, the proxy video and the camera path are rendered once for the whole video and remain consistent over its full length.
Following Code World Model~\cite{chen2026cwm}, the renderer processes them at inference time in overlapping 124-frame chunks, as illustrated in \cref{fig:supp_chunks}.
The last 34 generated frames of each chunk provide the continuation context of the next, so consecutive chunks join seamlessly.
For each chunk, we select the context views that best cover the part of the scene it traverses, so the renderer always sees the relevant appearance.

\vspace{1mm}\vspace{0pt}
\section{Dataset Curation}
\label{sec:supp_data}

\definecolor{palUnknown}{HTML}{30343B}
\definecolor{palObject}{HTML}{EEEEEE}
\definecolor{palVehicle}{HTML}{C66B63}
\definecolor{palFloor}{HTML}{707070}
\definecolor{palWall}{HTML}{A9B6C6}
\definecolor{palCeiling}{HTML}{BBA2DC}
\definecolor{palWindow}{HTML}{56C8D8}
\definecolor{palDoor}{HTML}{AD7756}
\definecolor{palHuman}{HTML}{F39442}
\definecolor{palSky}{HTML}{6495ED}
\definecolor{palWater}{HTML}{237DBE}
\definecolor{palTerrain}{HTML}{A58B68}
\definecolor{palVegetation}{HTML}{509650}
\definecolor{palBuilding}{HTML}{8798AD}
\definecolor{palInfra}{HTML}{EB871E}
\definecolor{palAnimal}{HTML}{CD964B}
\newcommand{\palswatch}[1]{{\setlength{\fboxsep}{0pt}\setlength{\fboxrule}{0.3pt}\fcolorbox{black!35}{#1}{\rule{0pt}{1.5ex}\rule{1.6em}{0pt}}}}
\newcommand{\palhex}[1]{{\scriptsize\texttt{#1}}}

\begin{table}[t]
    \centering
    \caption{\textbf{Category palette of the proxy video.} Each category is rendered in one fixed color. Ground covers both indoor floors and outdoor roads.}
    \label{tab:supp_palette}
    \footnotesize
    \setlength{\tabcolsep}{3pt}
    \renewcommand{\arraystretch}{1.15}
    \begin{tabular}{@{}cll@{\hspace{14pt}}cll@{}}
        \toprule
        \palswatch{palUnknown} & unknown & \palhex{30343B} & \palswatch{palDoor} & door & \palhex{AD7756} \\
        \palswatch{palObject} & object & \palhex{EEEEEE} & \palswatch{palSky} & sky & \palhex{6495ED} \\
        \palswatch{palHuman} & human & \palhex{F39442} & \palswatch{palWater} & water & \palhex{237DBE} \\
        \palswatch{palAnimal} & animal & \palhex{CD964B} & \palswatch{palTerrain} & terrain & \palhex{A58B68} \\
        \palswatch{palVehicle} & vehicle & \palhex{C66B63} & \palswatch{palVegetation} & vegetation & \palhex{509650} \\
        \palswatch{palFloor} & ground & \palhex{707070} & \palswatch{palBuilding} & building & \palhex{8798AD} \\
        \palswatch{palWall} & wall & \palhex{A9B6C6} & \palswatch{palInfra} & infrastructure & \palhex{EB871E} \\
        \palswatch{palCeiling} & ceiling & \palhex{BBA2DC} & \palswatch{palWindow} & window & \palhex{56C8D8} \\
        \bottomrule
    \end{tabular}
\end{table}

\paragraph{Training data.}
Training the renderer requires videos paired with frame-aligned proxy videos and context views of the same scene.
Such pairs do not need a complete 3D proxy world.
The renderer only sees the proxy through its frames, so any source from which a per-frame proxy can be rendered suffices, whether it provides 3D box annotations, tracked object poses, or simulator states.
We can thus combine sources that differ in scene type, camera motion, and interaction.
For real data, we convert annotations into proxy videos in the same style as at inference.
They cover indoor scenes from ARKitScenes~\cite{baruch2021arkitscenes} and ADT~\cite{pan2023adt}, driving scenes from Argoverse~\cite{chang2019argoverse}, hand-object interactions from HOT3D~\cite{banerjee2025hot3d} and DexYCB~\cite{chao2021dexycb}, and object-centric captures from Objectron~\cite{ahmadyan2021objectron}.
We also include the GTA~V gameplay data~\cite{chen2026cwm}, with its proxy videos mapped to our category palette.
We further render procedurally generated scenes from ProcTHOR~\cite{deitke2022procthor}, where we customize the camera trajectories, object interactions, and context views, insert new objects specified by a reference image, and allow object proxies with fine geometry.
In total, the training set contains about 40K pairs from eight sources.
These data are used exclusively for renderer training and are not involved in agentic reconstruction.

Each proxy frame is rendered from the annotations at the recorded camera of that frame.
Scene structure that the annotations lack is recovered from other ground truth of the dataset, \ie, the room layout from the scene mesh for ARKitScenes, surfaces from the depth for ADT, and the ground from the height map for Argoverse.
For ProcTHOR, the proxy is drawn from the simulator state, and we design clips with object interactions such as opening, throwing, and slicing, camera orbits, and bullet time.
Clips with several inserted objects further train the renderer to match each object to its own reference images.
The context views are other frames of the same recording, or frames of other cameras for Argoverse and DexYCB, and Qwen3-VL-8B~\cite{bai2025qwen3vl} writes the text prompt of each pair.

\paragraph{Proxy rendering.}
Training and inference proxies are rendered in exactly the same way, so the renderer sees the worlds built by the agent just as it saw the annotated data in training.
Each surface takes the flat color of its category in \cref{tab:supp_palette}, without lighting.
Colors encode categories rather than instances.
This allows a general proxy representation that specifies where things are and what they are, while their appearance comes from the context views.

\section{Evaluation Details}
\label{sec:supp_eval}

\subsection{Proxy World Reconstruction}
\label{sec:supp_eval_recon}

All methods take the same frames with the camera poses and depth maps provided by the datasets, and are scored against the same ground-truth boxes.

\paragraph{F1.}
F1 is computed at the scene level: the predicted and ground-truth oriented 3D boxes of each scene are matched one-to-one by 3D IoU, regardless of category, and a pair is a true positive when its IoU reaches the threshold, \ie, 0.25 or 0.5.
True positives, false positives, and false negatives are summed over each dataset, and F1 is the harmonic mean of precision and recall.
Following Boxer~\cite{detone2026boxer}, structural elements, \eg, walls, are ignored during evaluation.

\begin{table*}[t]
    \centering
    \caption{\textbf{Camera-controlled video generation with different numbers of context views.} All views are 9 on ScanNet++ and 8 on Tanks and Temples. TrajectoryCrafter and Holo4D operate on a source video and are not included.}
    \label{tab:supp_nviews}
    \small
    \setlength{\tabcolsep}{6pt}
    \begin{tabular}{lccccccccc}
        \toprule
        & \multicolumn{3}{c}{1 view} & \multicolumn{3}{c}{3 views} & \multicolumn{3}{c}{All views} \\
        \cmidrule(lr){2-4} \cmidrule(lr){5-7} \cmidrule(l){8-10}
        Method & PSNR $\uparrow$ & SSIM $\uparrow$ & LPIPS $\downarrow$ & PSNR $\uparrow$ & SSIM $\uparrow$ & LPIPS $\downarrow$ & PSNR $\uparrow$ & SSIM $\uparrow$ & LPIPS $\downarrow$ \\
        \midrule
        \multicolumn{10}{l}{\emph{ScanNet++~\cite{yeshwanth2023scannetpp}}} \\
        FramePack~\cite{zhang2025framepack} & 8.88 & 0.422 & 0.672 & 9.27 & 0.435 & 0.659 & 9.11 & 0.426 & 0.669 \\
        MiniMax H3~\cite{minimax2026h3} & 9.34 & 0.428 & 0.656 & 9.54 & 0.445 & 0.648 & 9.64 & 0.445 & 0.655 \\
        Code Video Model~\cite{liang2026codevideomodel} & 10.09 & 0.461 & 0.625 & 10.96 & 0.474 & 0.594 & 9.74 & 0.425 & 0.656 \\
        AnySplat~\cite{jiang2025anysplat} & 7.69 & 0.481 & 0.648 & 10.64 & 0.468 & 0.597 & 11.59 & 0.480 & 0.586 \\
        SEVA~\cite{zhou2025seva} & 10.98 & 0.496 & \tablesecond{0.580} & 14.24 & 0.586 & \tablesecond{0.445} & \tablesecond{15.15} & \tablesecond{0.606} & \tablesecond{0.418} \\
        Gen3C~\cite{ren2025gen3c} & \tablebest{11.85} & \tablesecond{0.501} & 0.582 & \tablesecond{14.91} & \tablesecond{0.606} & 0.481 & 14.74 & 0.603 & 0.486 \\
        FrameCrafter~\cite{wu2026framecrafter} & 9.67 & 0.322 & 0.705 & 12.46 & 0.490 & 0.552 & 14.35 & 0.587 & 0.459 \\
        \addlinespace[1pt]
        \textbf{\NICKNAME{} (Ours)} & \tablesecond{11.69} & \tablebest{0.529} & \tablebest{0.555} & \tablebest{15.52} & \tablebest{0.618} & \tablebest{0.425} & \tablebest{15.91} & \tablebest{0.622} & \tablebest{0.404} \\
        \midrule
        \multicolumn{10}{l}{\emph{Tanks and Temples~\cite{knapitsch2017tanks}}} \\
        FramePack~\cite{zhang2025framepack} & 10.70 & 0.231 & 0.652 & 10.53 & 0.228 & 0.656 & 10.65 & 0.234 & 0.645 \\
        MiniMax H3~\cite{minimax2026h3} & 11.34 & 0.273 & 0.640 & 11.47 & 0.265 & 0.635 & 11.58 & 0.265 & 0.632 \\
        Code Video Model~\cite{liang2026codevideomodel} & 11.60 & 0.271 & 0.623 & 11.80 & 0.278 & 0.619 & 11.68 & 0.267 & 0.625 \\
        AnySplat~\cite{jiang2025anysplat} & 6.03 & 0.200 & 0.698 & 8.29 & 0.217 & 0.651 & 11.17 & 0.266 & 0.595 \\
        SEVA~\cite{zhou2025seva} & 11.29 & 0.286 & 0.608 & 12.48 & 0.317 & 0.528 & 13.94 & 0.356 & 0.455 \\
        Gen3C~\cite{ren2025gen3c} & \tablesecond{12.87} & \tablesecond{0.323} & \tablebest{0.565} & \tablebest{14.03} & \tablebest{0.363} & \tablebest{0.494} & 14.51 & \tablesecond{0.383} & 0.481 \\
        FrameCrafter~\cite{wu2026framecrafter} & 10.37 & 0.209 & 0.689 & 13.70 & 0.334 & 0.524 & \tablesecond{14.91} & 0.375 & \tablebest{0.450} \\
        \addlinespace[1pt]
        \textbf{\NICKNAME{} (Ours)} & \tablebest{13.06} & \tablebest{0.324} & \tablesecond{0.568} & \tablesecond{13.89} & \tablesecond{0.350} & \tablesecond{0.521} & \tablebest{15.41} & \tablebest{0.384} & \tablesecond{0.451} \\
        \bottomrule
    \end{tabular}
\end{table*}

\subsection{Camera-Controlled Video Generation}
\label{sec:supp_eval_nvs}
All methods are given the same context views and the same target camera trajectory.
The frames along the trajectory are held out from every method, except for the first one, which serves as the starting view.

\paragraph{Baselines.}
Each baseline follows its native input setting.
SEVA~\cite{zhou2025seva}, Gen3C~\cite{ren2025gen3c}, and FrameCrafter~\cite{wu2026framecrafter} take the posed context views and the target cameras.
AnySplat~\cite{jiang2025anysplat} reconstructs the scene from the context views and renders it at the target cameras.
TrajectoryCrafter~\cite{yu2025trajectorycrafter} redirects the camera of a source video, so it starts from the first context view and follows the target cameras.
Holo4D~\cite{lan2026holo4d} also operates on a source video, and we provide it with a captured clip of the scene that is denser than our context views.
FramePack~\cite{zhang2025framepack} and MiniMax H3~\cite{minimax2026h3} accept no camera input, so they take the context views as visual references and the camera motion as text.
Code Video Model~\cite{liang2026codevideomodel} receives the same proxy video and context views as \NICKNAME.

\paragraph{PSNR, SSIM, and LPIPS.}
All frames are resized to $512\times384$, and each generated frame is paired with the captured frame at the same relative time along the trajectory, excluding the first frame.
We compute PSNR, SSIM, and LPIPS~\cite{zhang2018lpips} between each pair.
Scores are averaged over the frames of each scene and then across scenes.

\paragraph{FID.}
We pool the Inception features of all scored frames of a dataset and compare them with those of the corresponding captured frames, which gives one FID~\cite{heusel2017fid} per dataset.

\vspace{2mm}

\subsection{Physics-Grounded Video Generation}
\label{sec:supp_eval_physics}

All methods are given the same input view of each scenario, and all videos are cropped to it and resized to $512\times384$.

\paragraph{Baselines.}
The text-conditioned models, MiniMax H3~\cite{minimax2026h3} and Wan2.2 I2V-A14B~\cite{wan2025wan}, receive the interaction only through the text instruction of the scenario.
PhysGen3D~\cite{chen2025physgen3d} simulates the same objects with the same initial velocities, gravity, and friction in its own pipeline.
The controllable models are driven by signals rendered from the simulation of \NICKNAME: Wan2.2-Fun A14B~\cite{aigcapps2026videoxfun} by a depth video, Go-with-the-Flow~\cite{burgert2025gowiththeflow} by the optical flow of the moving objects, and Code Video Model~\cite{liang2026codevideomodel} by the proxy video itself.
These models thus share the physical outcome computed in our proxy world, and the comparison isolates how faithfully each of them renders it.

\paragraph{Trajectory IoU.}
We track every moving object in the generated video with SAM~2~\cite{ravi2025sam2}, prompted by its simulated box in the first frame, and compare the tracked mask with the simulated one every $1/8$\,s.
We take the IoU between the bounding boxes of the two masks, averaged over the objects and time steps of a scenario and then across scenarios.

\paragraph{Trajectory error.}
Trajectory error is the pixel distance between the displacements of the tracked and simulated mask centroids from the first frame, averaged the same way.

\paragraph{VLM scores.}
Qwen3-VL-32B~\cite{bai2025qwen3vl} rates physical realism, photorealism, and semantic consistency in $[0,1]$ with the prompt below, which adapts the criteria of PhysGen3D~\cite{chen2025physgen3d}.
The judge sees the input image, the instruction of the scenario, and ten frames evenly spaced over the interaction, which are taken at the same instants for all methods.
Scores are averaged across scenarios.

\begin{promptbox}{Prompt of the VLM judge}
\begin{Verbatim}
You are evaluating a generated video. The first image is the original input image. The next 10 images are frames evenly sampled, in temporal order, from the first <window> seconds of the generated video. The video should follow this instruction: "<instruction>"

Evaluate the video on three criteria, each scored from 0 to 1 (1 is best):
- Physical realism: how realistically the video follows physical rules and whether it represents real physical properties such as gravity, inertia, collision response, elasticity and friction. Penalize videos in which the described motion does not happen or the scene stays completely still.
- Photorealism: the overall visual quality of the video, including visual artifacts, discontinuities, objects that deform, duplicate or disappear, and how accurately it replicates light, shadow, texture and materials.
- Semantic consistency: how well the content of the video matches the intended meaning of the instruction, while staying the same scene with the same objects as the input image.

First write the concrete evidence you see in the frames for each criterion (what moves, how it moves, any violation or artifact). Then give the scores. Answer only with JSON:
{"evidence": {"physical_realism": "...", "photorealism": "...", "semantic_consistency": "..."}, "physical_realism": <0-1>, "photorealism": <0-1>, "semantic_consistency": <0-1>}
\end{Verbatim}
\end{promptbox}

\section{Additional Results}
\label{sec:supp_comp}

\begin{figure*}[t]
    \centering
    \includegraphics[width=\textwidth]{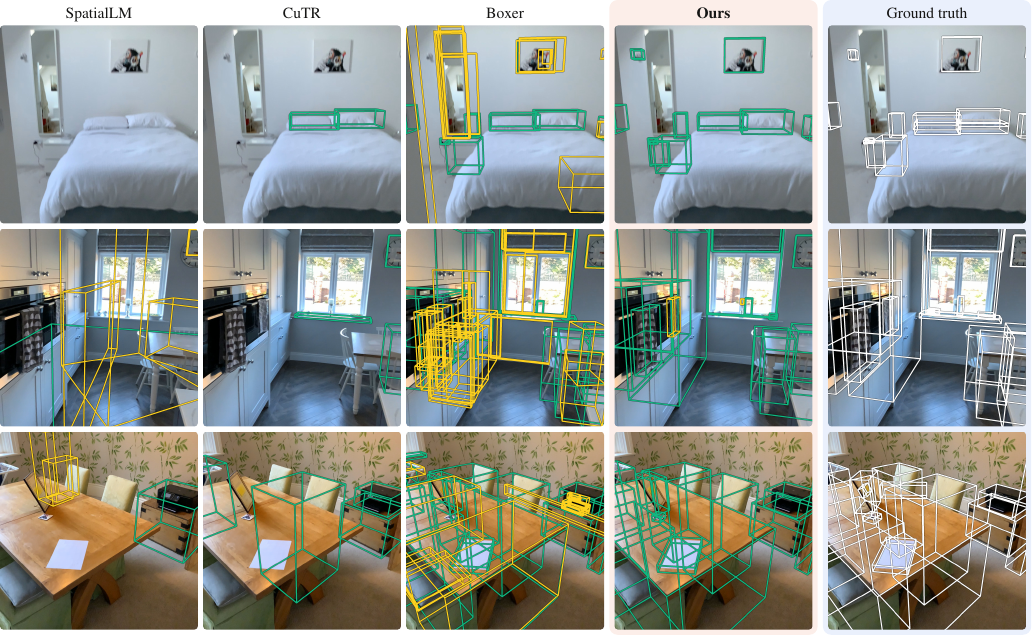}
    \caption{\textbf{Qualitative comparison of 3D proxy world reconstruction} with SpatialLM~\cite{mao2025spatiallm}, CuTR~\cite{lazarow2025cubify}, and Boxer~\cite{detone2026boxer} on three \mbox{CA-1M}~\cite{lazarow2025cubify} scenes. The 3D boxes of each method are projected onto the same input frame. For all methods, predicted boxes matched to the ground truth at IoU 0.25 are drawn in green, and duplicate or spurious boxes that match no ground-truth box are drawn in yellow. Ground-truth boxes are drawn in white. SpatialLM and CuTR miss most objects and Boxer adds many duplicate and spurious boxes, whereas \NICKNAME{} stays close to the ground truth.}
    \label{fig:supp_recon}
\end{figure*}

\begin{figure*}[p]
    \centering
    \includegraphics[width=\textwidth]{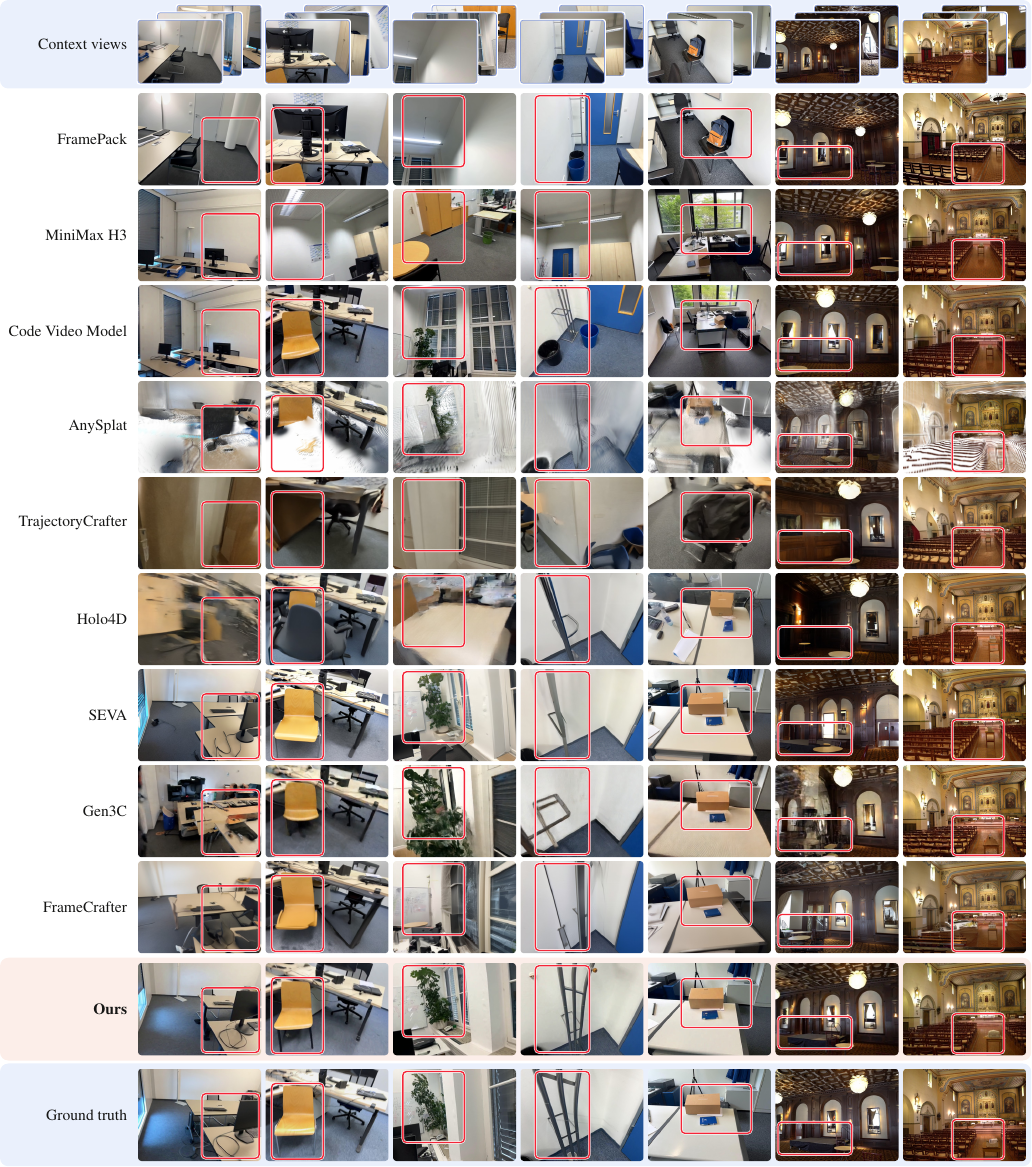}
    \caption{\textbf{More qualitative comparisons of camera-controlled video generation} with all baselines. Each column shows one target frame of a test scene, with its context views at the top. As outlined in red, the baselines miss the target viewpoint, blur, or hallucinate content, whereas \NICKNAME{} stays faithful to the ground truth. Zoom in for the best view.}
    \label{fig:supp_nvs}
\end{figure*}

\begin{figure*}[t]
    \centering
    \includegraphics[width=\textwidth]{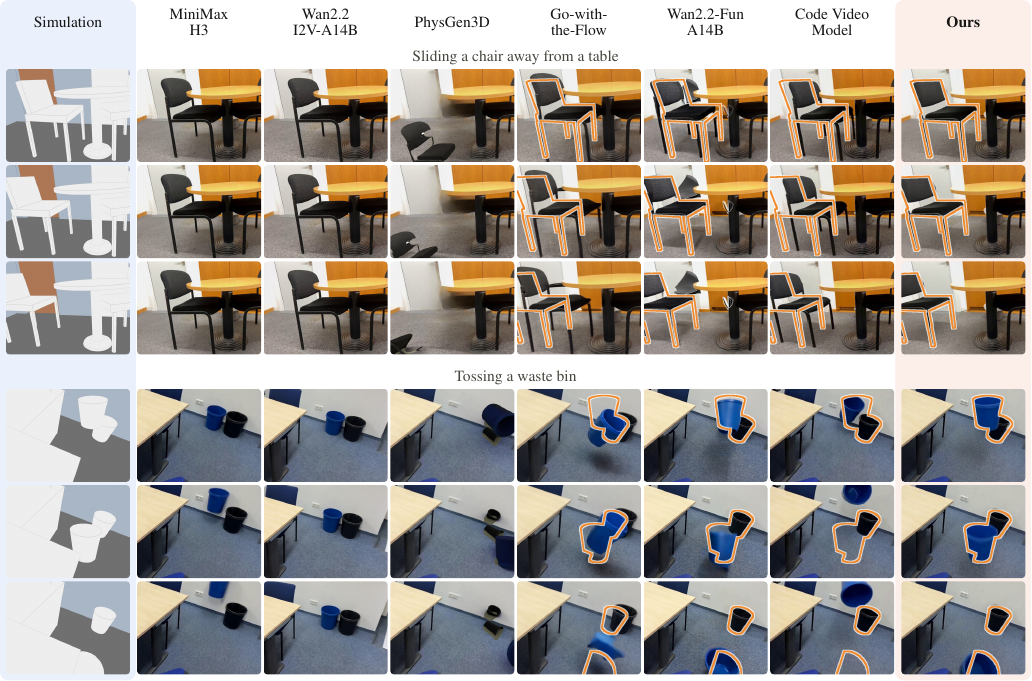}
    \caption{\textbf{More qualitative comparisons of physics-grounded video generation} with MiniMax H3~\cite{minimax2026h3}, Wan2.2 I2V-A14B~\cite{wan2025wan}, PhysGen3D~\cite{chen2025physgen3d}, Go-with-the-Flow~\cite{burgert2025gowiththeflow}, Wan2.2-Fun A14B~\cite{aigcapps2026videoxfun}, and Code Video Model~\cite{liang2026codevideomodel} on two ScanNet++ scenes. The first column shows the simulation as the reference, and the three rows of each scene are consecutive time steps from top to bottom. For the methods driven by our simulation, the simulated objects are outlined in orange. The baselines leave the object in place, move it the wrong way, or deform and duplicate it, whereas \NICKNAME{} follows the simulated motion and keeps the shape of the objects.}
    \label{fig:supp_physics}
\end{figure*}

\begin{figure*}[t]
    \centering
    \includegraphics[width=\textwidth]{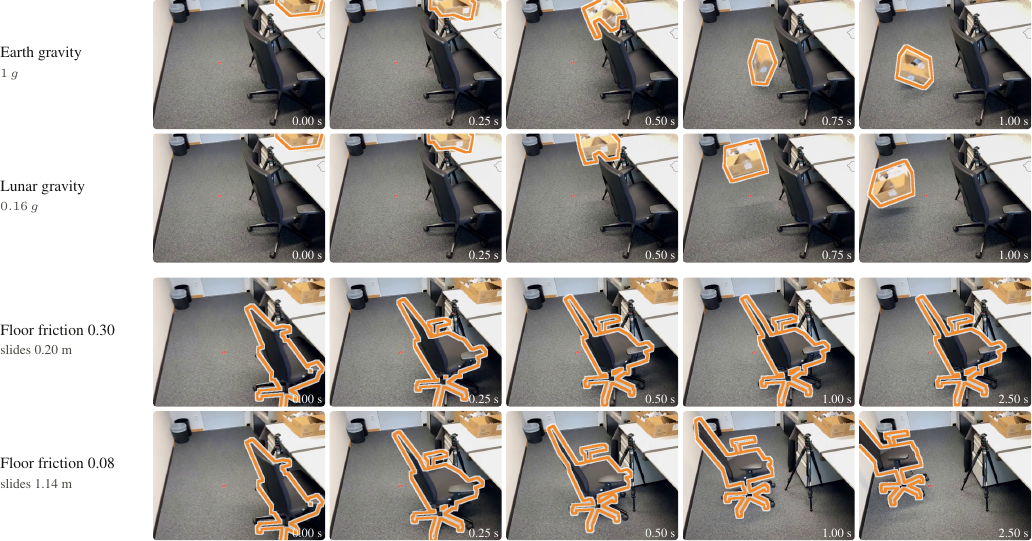}
    \caption{\textbf{Control over physical properties.} Each pair of rows changes a single physical property in the scene program while the scene and the action stay fixed, with the simulated object outlined in orange. Top: under lunar gravity, the box pushed off the desk is still in the air at 1.00\,s, when it already rests on the floor under Earth gravity. Bottom: lowering the floor friction from 0.30 to 0.08 lets the pushed chair slide much farther. \NICKNAME{} faithfully follows the changed dynamics in both cases.}
    \label{fig:supp_physics_sweep}
\end{figure*}
\begin{figure*}[t]
    \centering
    \includegraphics[width=\textwidth]{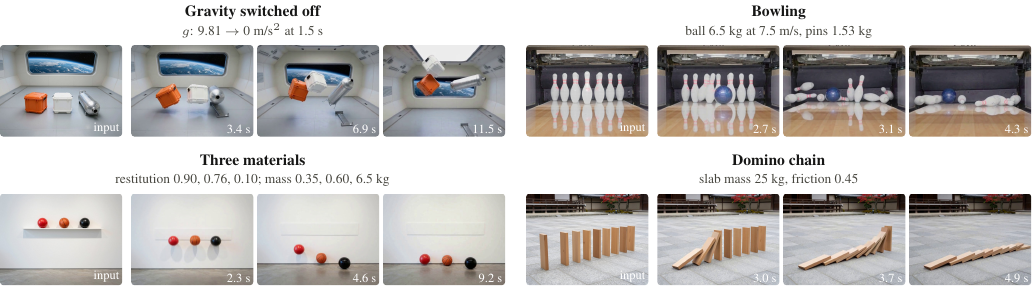}
    \caption{\textbf{Physical settings in worlds built from a single image.} Each case shows the input image and three instants of the generated video, with the settings of its scene program above. Global settings can change during the video, objects of the same shape can differ in material, and objects that are not in the image can be added, \eg, the bowling ball. In the three-materials case, restitution and mass are listed from left to right. The red rubber ball bounces highest, and the heavy black ball barely leaves the floor.}
    \label{fig:supp_physics_image}
\end{figure*}

\paragraph{Number of context views.}
We further evaluate camera-controlled video generation with different numbers of context views.
In \cref{tab:supp_nviews}, all methods receive the same 1, 3, or all context views of each test scene, \ie, 9 views on ScanNet++ and 8 on Tanks and Temples.
Different from the ablation in \cref{tab:rendering_ablation} of the main paper, where only the views given to the renderer are reduced, here \NICKNAME{} uses the given views alone for both reconstruction and rendering.
\NICKNAME{} achieves the best or comparable results under all settings, from a single image to all views, and improves steadily as more views are given.
This shows that the agent builds a reliable proxy world even from sparse observations, and that the renderer benefits from every additional view.

\paragraph{3D proxy world reconstruction.}
In \cref{fig:supp_recon}, we provide qualitative comparisons of 3D proxy world reconstruction, where the reconstructed boxes are projected onto the input frames.
SpatialLM and CuTR detect only a few of the objects, and Boxer finds most of them but buries them under duplicate and spurious boxes.
Thanks to the spatial reasoning of the agent, \NICKNAME{} keeps one box per object and removes the spurious ones, giving the reconstruction closest to the ground truth.

\paragraph{Camera-controlled video generation.}
In \cref{fig:supp_nvs}, we provide more qualitative comparisons of camera-controlled video generation with all baselines.
Beyond the quantitative results in \cref{tab:nvs} of the main paper, the visual differences between the methods are clear.
Gen3C distorts the geometry and leaves ghosting artifacts once the camera moves away from the context views, \eg, the desk in the first scene and the coat stand in the fourth.
SEVA and FrameCrafter produce sharp frames that are inconsistent with the scene: objects such as the monitor and the plant go missing, and a doorway or a fireplace is hallucinated in the Ballroom.
The other baselines miss the target viewpoint or degrade into blur and floaters.
In contrast, \NICKNAME{} reaches the target viewpoint and preserves the objects of the scene with their appearance, since the proxy video fixes the layout in 3D.

\vspace{1mm}

\paragraph{Physics-grounded video generation.}
In \cref{fig:supp_physics}, we provide more visual comparisons with all baselines.
The text-conditioned models barely move the objects.
PhysGen3D and Go-with-the-Flow deform the objects, Wan2.2-Fun follows the simulated depth but adds a second backrest to the chair and flattens the bin, and Code Video Model duplicates the chair.
In contrast, \NICKNAME{} follows the simulation in both interactions while keeping the objects intact.
Beyond following a given simulation, \NICKNAME{} allows users to control the physical properties of the world and thereby simulate different interactions in the same scene.
In \cref{fig:supp_physics_sweep}, we show examples in which a single physical property in the scene program is changed, \ie, the gravity or the friction of the floor, while the scene and the action stay fixed.
Under lunar gravity, the box pushed off the desk is still falling when it already rests on the floor under Earth gravity, and lowering the floor friction from 0.30 to 0.08 lets the chair slide 1.14\,m instead of 0.20\,m.
In \cref{fig:supp_physics_image}, the same control applies to worlds built from a single image, where global settings can change over time and objects of the same shape can differ in material.

\section{More Applications}
\label{sec:supp_apps}

Every video of \NICKNAME{} is backed by an explicit and editable 3D proxy world.
In this section, we show several applications that this enables.

\begin{figure*}[t]
    \centering
    \includegraphics[width=\textwidth]{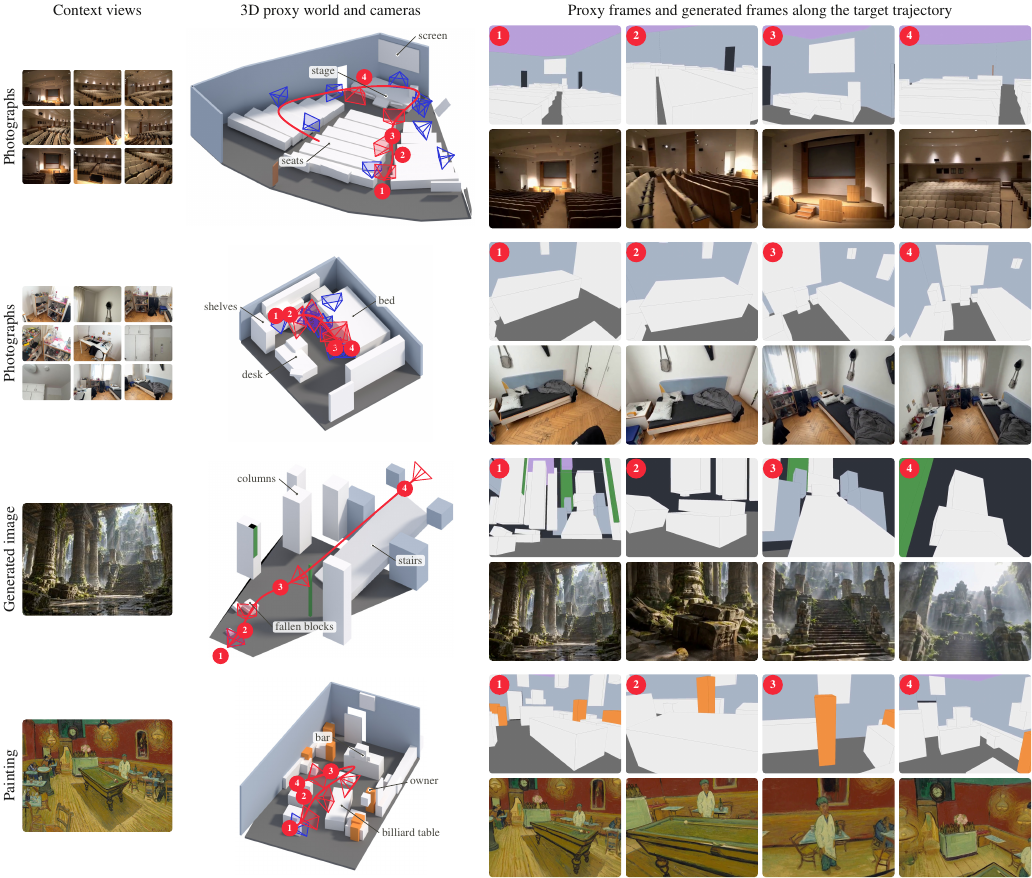}
    \caption{\textbf{Spatial exploration.} Each row shows the context views, the 3D proxy world with the context cameras (blue) and the target camera trajectory (red), and the proxy frames (top) and generated frames (bottom) at the four numbered target cameras. For some scenes, the ceiling and the walls facing the viewer are hidden to reveal the interior structure. From top to bottom: nine photographs of the Auditorium scene of Tanks and Temples~\cite{knapitsch2017tanks}, nine photographs of a bedroom, a single generated image, and Vincent van Gogh, \emph{The Night Caf\'e}, 1888. Across diverse scenes and camera trajectories, the generated frames follow the layout of the proxy frames while keeping the appearance of the context views, also for views far from them.}
    \label{fig:supp_explore}
\end{figure*}
\begin{figure*}[tp]
    \centering
    \includegraphics[width=\textwidth]{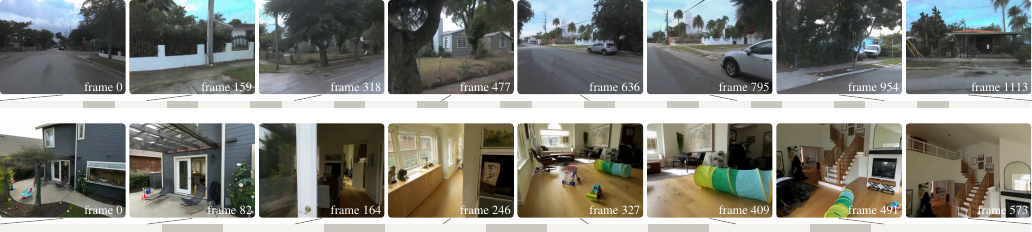}
    \caption{\textbf{Long takes.} Top: 1114 frames along a street reconstructed from driving captures. Bottom: 574 frames from the garden into the living room of a house reconstructed from photographs. The bar under each row is the time axis of the take, the darker segments mark the frames shared by consecutive 124-frame chunks, and the thin lines point to the shown frames. The scene stays consistent across chunks.}
    \label{fig:supp_longtake}
\end{figure*}
\paragraph{Spatial exploration.}
\NICKNAME{} turns a few images into a world that can be explored freely.
As shown in \cref{fig:supp_explore}, the input can be as diverse as a handful of photographs of a real place, a single image, or even a painting.
Along diverse camera trajectories, \NICKNAME{} synthesizes novel views far beyond the context views, with a consistent layout and a faithful appearance.
With the long video inference of \cref{sec:supp_impl_long}, \NICKNAME{} can support long takes of over a thousand frames through an entire scene, as shown in \cref{fig:supp_longtake}.

\begin{figure*}[tp]
    \centering
    \includegraphics[width=\textwidth]{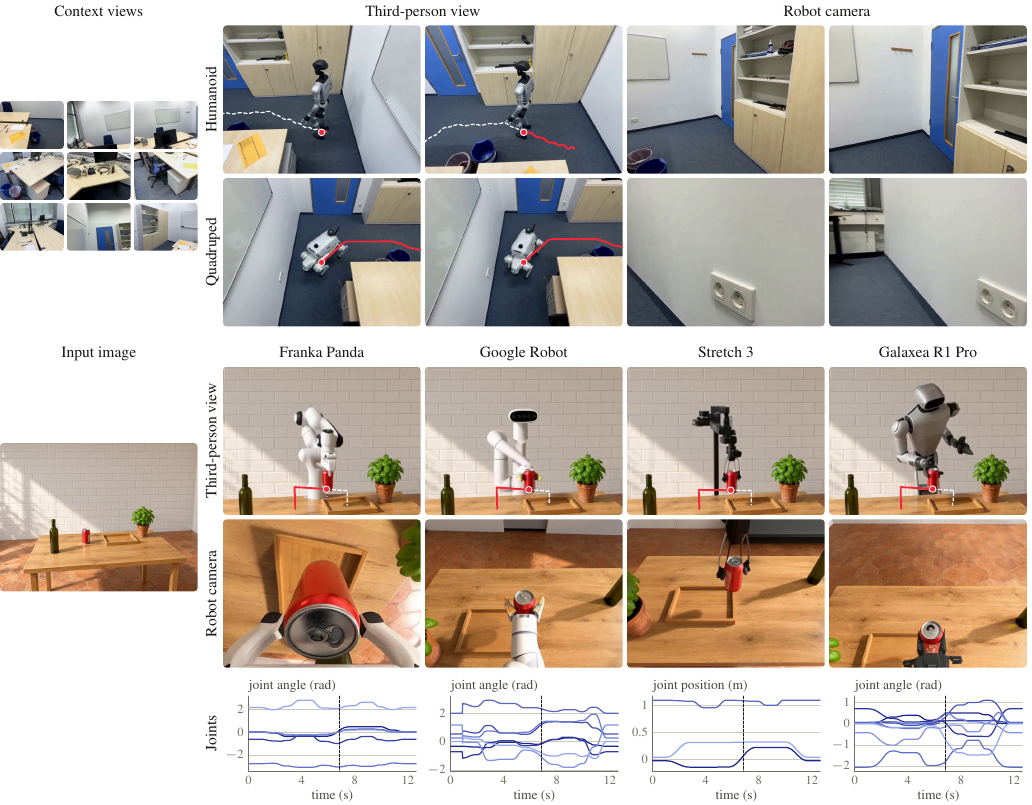}
    \caption{\textbf{Embodied simulation.} Simulation-ready assets, \eg, the humanoid and the manipulators, can be placed into the 3D proxy worlds reconstructed by \NICKNAME{}, simulated, and rendered into videos. Top: a humanoid and a quadruped navigate an office reconstructed from nine context views, following routes specified in the proxy world, which the humanoid tracks with a pretrained walking policy~\cite{unitree2024rlgym}. Bottom: four robots place the same can on the tray in a world built from the input image, with trajectories solved by inverse kinematics and the grasp resolved by contact simulation. Each execution is rendered from a third-person view and from the robot's own camera, and comes with its simulation state: the path of the robot base (top) and of the can (bottom) is overlaid on the third-person views, and the last row plots the joint trajectories of each manipulator, with a dashed line at the shown instant.}
    \label{fig:supp_robots}
\end{figure*}

\vspace{1mm}

\paragraph{Embodied simulation.}
Building on its support for dynamic simulation, \NICKNAME{} can serve as a simulator for embodied agents.
We take embodiments, \eg, simulation-ready robots from existing asset libraries, and place them into the 3D proxy world that \NICKNAME{} reconstructs for a specific scene.
In \cref{fig:supp_robots}, a humanoid and a quadruped navigate an office, and four robots place the same can on a tray.
Different embodiments can thus be evaluated on the same task under the same specified scene and physical conditions.
\NICKNAME{} renders each execution from a third-person view and from the robot's head or wrist camera.
The task trajectories and physical interaction states accompany the rendered observations, providing data for robot training and for comparing behaviors before real deployment.

\begin{figure*}[tp]
    \centering
    \includegraphics[width=\textwidth]{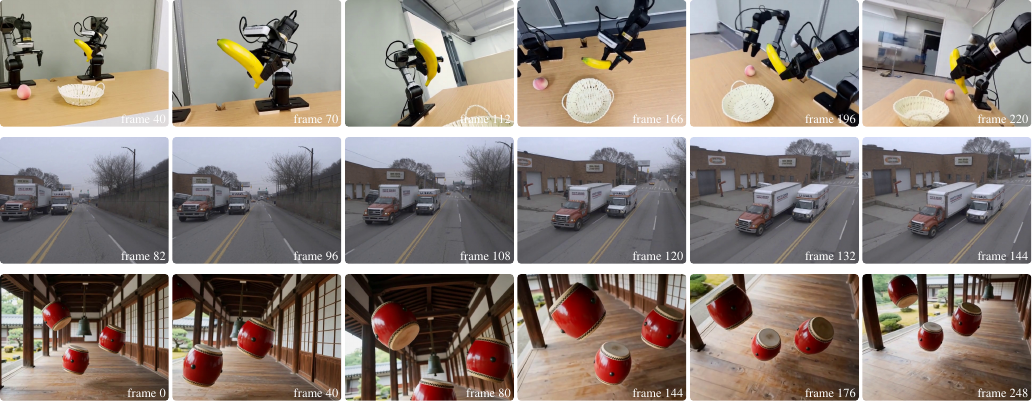}
    \caption{\textbf{Bullet time.} Time is frozen at one instant while the camera flies through the scene. Top: two robot arms captured by four synchronized phones. Middle: a street with oncoming trucks, captured by the cameras of a driving car. Bottom: falling drums simulated in a world built from a single generated image. The frozen objects stay in place and keep their appearance far from the input views.}
    \label{fig:supp_bullettime}
\end{figure*}
\paragraph{Bullet time.}
Bullet time freezes a moment while the camera keeps moving through the scene.
Given a dynamic scene captured by a few synchronized cameras, \eg, four phones at different viewpoints, \NICKNAME{} reconstructs the proxy world at a single instant and moves the camera freely with this static state fixed.
In \cref{fig:supp_bullettime}, we show bullet-time results, where each frozen moment is observed from different viewpoints.
Please see the videos on our project page.

\begin{figure*}[tp]
    \centering
    \includegraphics[width=\textwidth]{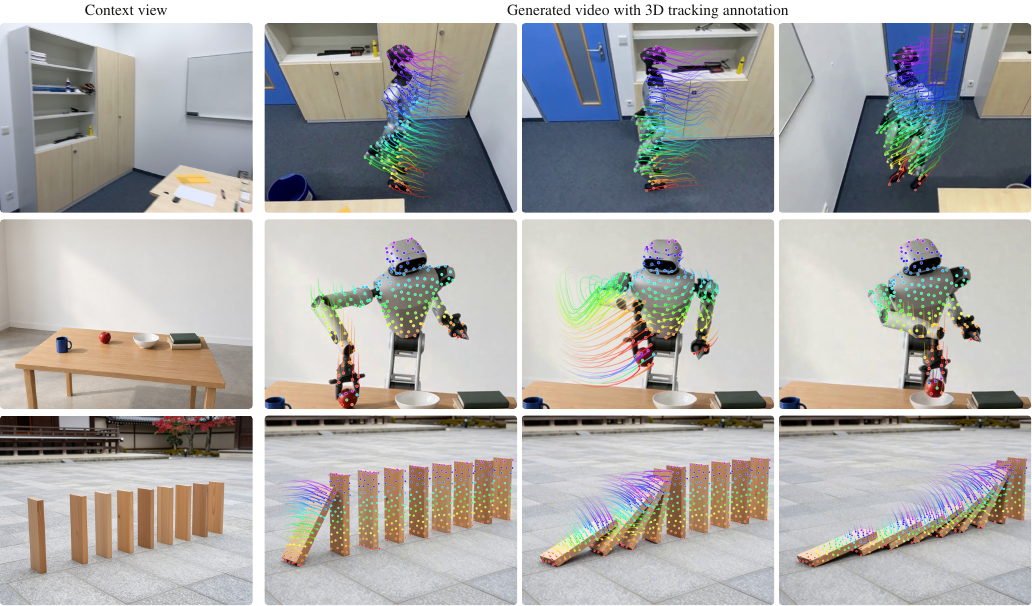}
    \caption{\textbf{Dense 3D tracking.} \NICKNAME{} generates each video from a simulated world state, so the 3D trajectory of every point on the simulated objects comes with the video.}
    \label{fig:supp_tracking}
\end{figure*}
\paragraph{Dense 3D tracking.}
\NICKNAME{} renders every video from an explicit world state, so each point on the simulated objects comes with its 3D trajectory without running a tracker, as shown in \cref{fig:supp_tracking}.
As scenes and interactions can be customized, this offers a scalable source of diverse, long-tail training data with dense annotations for 4D reconstruction and 3D tracking methods~\cite{feng2025st4rtrack,liu2026traceanything,karhade2026any4d,luo20264rc,sucar2026vdpm,xiao2025spatialtrackerv2}.

\section{Video Results}
\label{sec:supp_video}
\vspace{5pt}

As still frames cannot fully convey camera motion, physical dynamics, and temporal consistency, we provide video results on our project page, \url{https://mundo-world-model.github.io/}, which includes more visual results.
We recommend watching the video results.

\section{Limitations and Future Work}
\label{sec:supp_limit}
\vspace{-1pt}

Overall, \NICKNAME{} pushes the boundary of customizable world models and lays out a promising route.
It still has several limitations, as discussed below, which also point to directions for future work.

\paragraph{Rigid bodies and assumed parameters.}
In the current version, \NICKNAME{} simulates objects as rigid bodies.
Soft and deformable objects, \eg, cloth and fluids, remain hard to model and simulate precisely.
Mass, friction, and restitution are not observed in the context views, so the agent sets plausible values.
Estimating them from observed motion in future work would make the system more complete.

\paragraph{Fine-grained interaction.}
Small objects are hard to model precisely: the reconstructed points contain floaters, and the boxes are hard to align exactly.
The renderer is also prone to hallucination in interactions that involve handling small objects, \eg, dexterous manipulation.

\paragraph{Long-horizon memory.}
\NICKNAME{} has no long-horizon memory mechanism yet.
Although the context views preserve the static scene, details may change after extended dynamic changes or when a previously occluded or unseen region comes into view.
Equipping the renderer with a memory of the generated content is left for future work.

\paragraph{Efficiency.}
\NICKNAME{} does not run in real time in its current form.
Reconstructing a scene takes tens of minutes, and the renderer runs well below real time.
We have not yet accelerated the inference of the video model.
Few-step distillation and streaming inference, \eg, StreamRender-H3~\cite{lu2026streamrenderh3}, are promising ways toward interactive rendering.

\end{document}